\documentclass[acmlarge,screen]{acmart}
\usepackage[utf8]{inputenc}
\usepackage[T1]{fontenc}

\AtBeginDocument{%
  \providecommand\BibTeX{{%
    \normalfont B\kern-0.5em{\scshape i\kern-0.25em b}\kern-0.8em\TeX}}}

\setcopyright{none}

\newcommand{\etal}{\textit{et al.}}

\begin{document}

\title[Providing insight and minimizing false negative rate in melanoma detection]{Minimizing false negative rate in melanoma detection and providing insight into the causes of classification}


\author{Ellák Somfai}
\email{somfaiellak@inf.elte.hu}
\orcid{0000-0002-2218-8855}
\affiliation{\institution{Eötvös University} \city{Budapest} \country{Hungary}}
\affiliation{\institution{Wigner Research Centre for Physics} \city{Budapest} \country{Hungary}}
\authornote{corresponding author}

\author{Benjámin Baffy}
\affiliation{\institution{Eötvös University} \city{Budapest} \country{Hungary}}

\author{Kristian Fenech}
\affiliation{\institution{Eötvös University} \city{Budapest} \country{Hungary}}

\author{Changlu Guo}
\affiliation{\institution{Eötvös University} \city{Budapest} \country{Hungary}}

\author{Rita Hosszú}
\affiliation{\institution{Semmelweis University} \city{Budapest} \country{Hungary}}

\author{Dorina Korózs}
\affiliation{\institution{Semmelweis University} \city{Budapest} \country{Hungary}}

\author{Fabrizio Nunnari}
\affiliation{\institution{German Research Center for Artificial Intelligence} \city{Saarbrücken} \country{Germany}}

\author{Marcell Pólik}
\affiliation{\institution{Eötvös University} \city{Budapest} \country{Hungary}}

\author{Daniel Sonntag}
\affiliation{\institution{German Research Center for Artificial Intelligence} \city{Saarbrücken} \country{Germany}}

\author{Attila Ulbert}
\affiliation{\institution{Eötvös University} \city{Budapest} \country{Hungary}}

\author{András Lőrincz}
\email{lorincz@inf.elte.hu}
\orcid{0000-0002-1280-3447}
\affiliation{\institution{Eötvös University} \city{Budapest} \country{Hungary}}

%
%
%

\renewcommand{\shortauthors}{Somfai, et al.}

\begin{abstract}
Our goal is to bridge human and machine intelligence in melanoma detection. We develop a classification system exploiting a combination of visual pre-processing, deep learning, and ensembling for providing explanations to experts and to minimize false negative rate while maintaining high accuracy in melanoma detection. Source images are first automatically segmented using a U-net CNN. The result of the segmentation is then used to extract image sub-areas and specific parameters relevant in human evaluation, namely center, border, and asymmetry measures. These data are then processed by tailored neural networks which include structure searching algorithms. Partial results are then ensembled by a committee machine. Our evaluation on the largest skin lesion dataset which is publicly available today, ISIC-2019, shows improvement in all evaluated metrics over a baseline using the original images only. We also showed that indicative scores computed by the feature classifiers can provide useful insight into the various features on which the decision can be based.
\end{abstract}



\maketitle

\section{Introduction}

An important observation in large scale image classification and other image processing tasks is that custom feature engineering is detrimental and the best approach is to feed the model the original data without any loss of information. However, deep networks lack the explanatory power about their outputs and in case of limited datasets, such as for skin lesions where the typical dataset sizes range from few hundreds to few ten thousands. The insight collected by human experts can significantly improve the performance of machine learning applications as well, as we demonstrate in this paper.

Dermoscopic imaging has been the prevalent method of diagnosing melanoma since the 1990s~\cite{rigel2010evolution}. Clinical diagnosis can be made by specialists assessing dermoscopic images against checklists of common attributes, described in methods such as ABCDE checklist~\cite{rigel2005abcde}, Menzies criteria~\cite{menzies1996frequency} or the 7-point method~\cite{mackie1990clinical}. These methods are not fool-proof as there exist cases of melanoma which do not fit into the prescribed descriptions~\cite{pizzichetta2007dermoscopic}. 

The task of melanoma diagnosis is difficult due to the existence of a variety of benign nevi, which while generally distinguishable based on the diagnostic criteria for melanoma, some atypical nevi can more closely resemble their malignant counterpart~\cite{grant1999misdiagnosis}, providing opportunities for false negative classification of melanoma as an atypical nevus.

In skin lesion classification, Esteva \etal\  \cite{esteva_dermatologist-level_2017} demonstrated that the performance of deep learning methods can be comparable to dermatologists. See also \cite{brinker_deep_2019} for more recent results. However, deep learning models are also prone to confusion between malignant melanoma and benign nevi. The origin of this confusion stems in part from the fact that the melanocytes (the cells producing brown pigment in the skin), which form benign clusters in nevi, can grow out of control in the malignant melanoma. Hence, there is room for improvements.

There are a number of publicly available skin lesion datasets \cite{mendonca2013ph2,kawahara2019derm7pt}, of which the most comprehensive and largest is associated with the International Skin Imaging Collaboration (ISIC).  The version of the ISIC dataset, which was made available as part of the 2019 skin lesion challenge \cite{tschandl_ham10000_2018, codella_skin_2017, combalia_bcn20000:_2019}, contains over 25,000 images across 8 diagnostic classes. However, this dataset presents a number of difficulties for deep learning. The dataset itself is heavily imbalanced, with nevi making up just over 50\% of the images. Image quality and resolution is highly variable. This presents a challenge as many key attributes of melanoma exist on a much smaller length scale than the lesion itself, therefore the presence of blur, insufficient magnification or poor image resolution can make these features unresolvable to the model being trained.

Our work is based on the observation that texture, shape and size variations of skin lesions are very large compared to other image domains e.g., those of faces. In addition, subtleties, such as pseudo pods, the presence or absence of sharp edges at the border, barely visible peppering-like, white-bluish structures among many others can be relevant, although fine signs of serious skin problems. This high variety may pose serious demand on the size of the database, especially for rare cases, if deep learning methods are to be exploited for classification. Instead, we suggest to develop a few filters that try to enhance the specific features at least implicitly. We use sub-networks. Each of them was trained on a specific representation of the lesion such as detailed internal structure, border, asymmetry in color, presence of blue-whitish structures (often present in melanoma) as well as the regular structure extracting method. Although such enhancing filters are at a preliminary design phase, we found that they can help in the classification by ensembling them via a shallow fully-connected neural network which provides the final class prediction.

An additional advantage of our approach is that it can provide explanation to the expert about the details of the classification in terms of the features trained separately. We shall elaborate on this in Section~\ref{sec:results-indications}. The paper is organized as follows. The next section (Sect.~\ref{sec:relworks}) reviews the related works. It is followed by the applied methods (Sec.~\ref{sec:methods}. Results and discussions can be found in Sect.~\ref{sec:res-disc}. The paper ends with a short conclusion about the applicability of our methods and an outlook to future options.

\section{Related work}\label{sec:relworks}

There has been significant work on the application of pure deep convolutional neural networks for the detection of melanomas~\cite{codella_skin_2017,kawahara_deep_2016}. Brinker \etal~\cite{brinker_deep_2019} demonstrated that a deep convolutional neural network, specifically a ResNet50 architecture, can outperform dermatologists in classifying melanoma and atypical nevi from dermoscopic images. 

For work based on the ISIC 2019 dataset, the top-performers of the classification challenge of 2019 were all based on deep learning and in some cases ensembles of deep neural networks. Gessert \etal~\cite{gessert2019skin} trained multiple variants of different deep convolutional networks, predominately using EfficientNet architectures. These networks are trained at different resolutions and the final prediction is made by a weighted sum over the classifiers. In this case no special attention is given to particular features of the images outside of cropping the lesion.

Prior to the surge in deep learning based methods, work focusing on the classification of melanoma and nevi using conventional statistical learning methods have been explored. In the work of Rastgoo \etal~\cite{rastgoo:hal-01457799} the task of distinguishing melanoma and nevi is achieved by performing local feature extraction which takes into account such attributes as lesion shape, border asymmetry, colour variation and texture variation. In this work the authors used their extracted features with three classifiers: support vector machine, random forest and gradient boosting. It was shown that the random forest classifier provided the highest performance with their extracted features.

In the present work the above two approaches are combined: we exploit the human effort that went into identifying decisive features, and harness the power of deep convolutional networks.

\section{Methods}\label{sec:methods}

Our approach to assemble a skin lesion classifier is to build a number of deep CNN-based classifiers, which we call ``feature classifiers'', which receive preprocessed input images typically focused on the individual criteria of the ABCD melanoma classification rules, and fuse their predictions via a shallow neural net. Our feature classifiers work either on the whole input image, or typically use a preprosessed image of the lesion. 

\begin{figure}[t]
    \centering
    \includegraphics[width=0.65\textwidth]{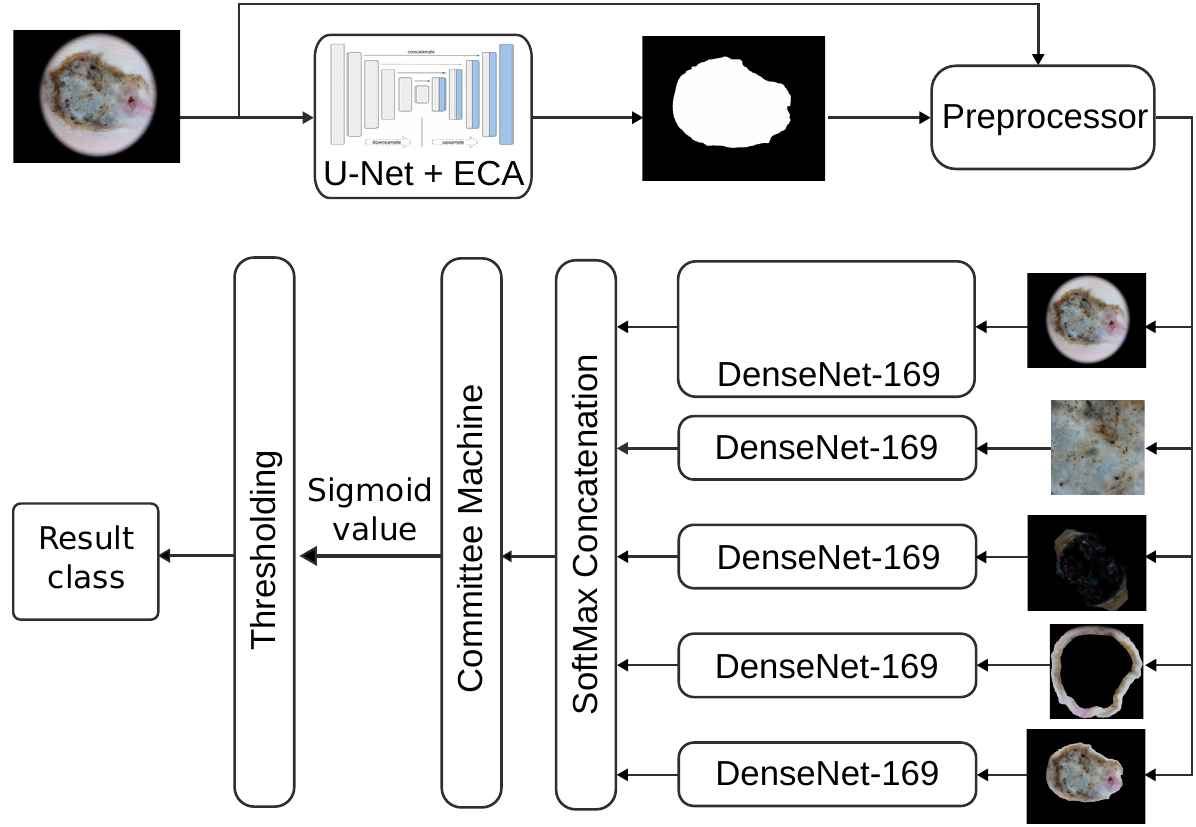}
    \caption{\label{fig:method-overview}
        Overview of the architecture of our method. The feature classifiers (based on DenseNet) operate on feature-highlighted input; their results are combined by a shallow net committee machine.
        }
\end{figure}

Figure \ref{fig:method-overview} shows an overview of the classification method. 
The input image is first segmented by a U-net based model (Section \ref{sec:segmenter}) to obtain the mask of the lesion. The mask is used to preprocess the original images to highlight the various features; these are fed into the feature classifiers (Section \ref{sec:feature-classifiers}). The ensemble of the feature classifiers' output (either the combined softmax layers, or the previous dense feature layer) is inputted into a ``committee machine'' --- a shallow neural net responsible for a learned fusion of the individual predictions (Section \ref{sec:committee-machine}) --- which yields the final classification.

\subsection{Lesion segmenter}
\label{sec:segmenter}

Many of our feature classifiers depend on the image mask of the lesion. The required foreground-background segmentation is achieved by a U-net-based convnet\cite{navab_u-net:_2015}, with modified skip connections.

\begin{figure}
    \centering
    \includegraphics[width=0.75\textwidth]{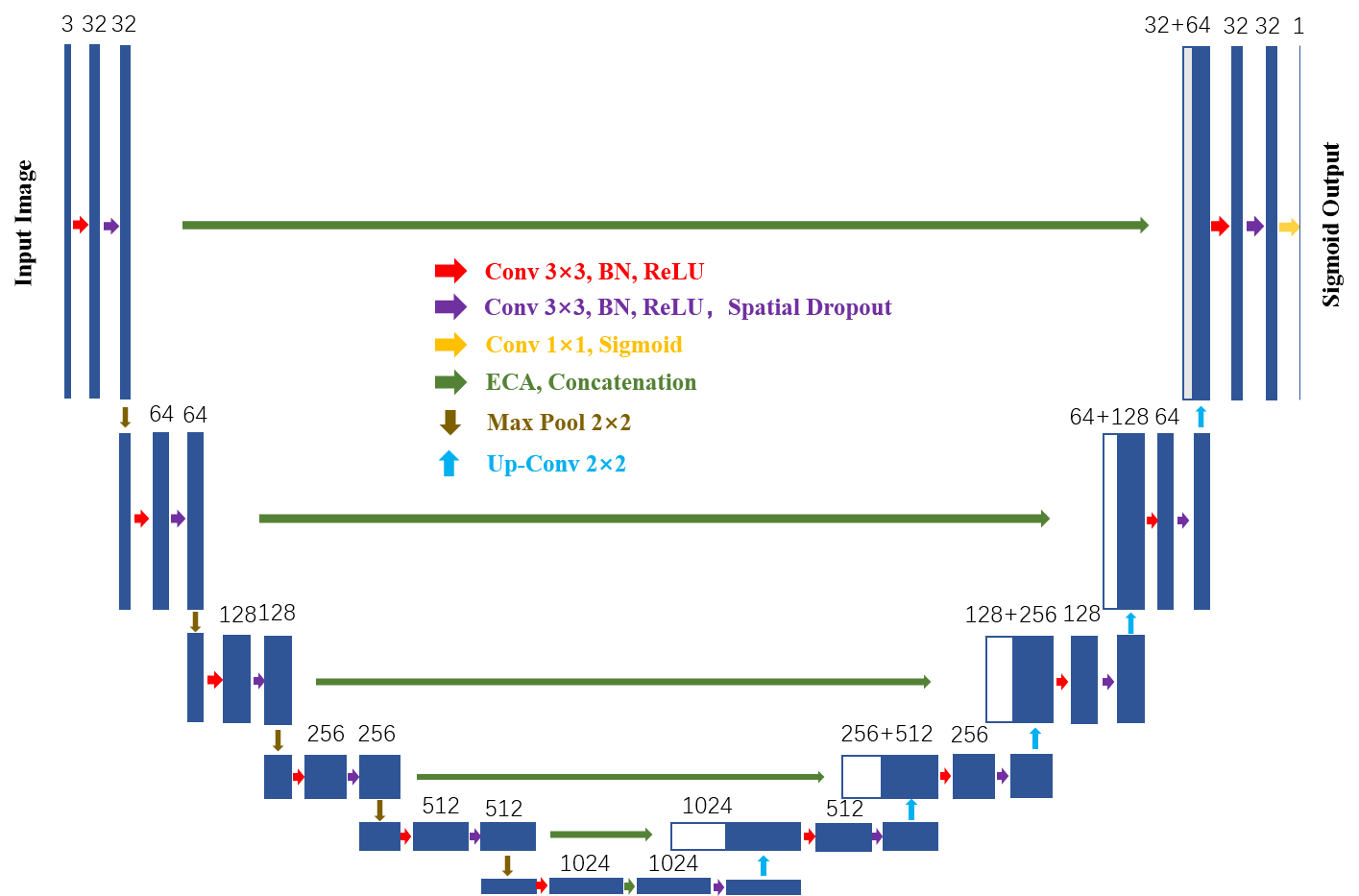}
    \vspace*{5mm}
    
    \includegraphics[width=0.85\textwidth]{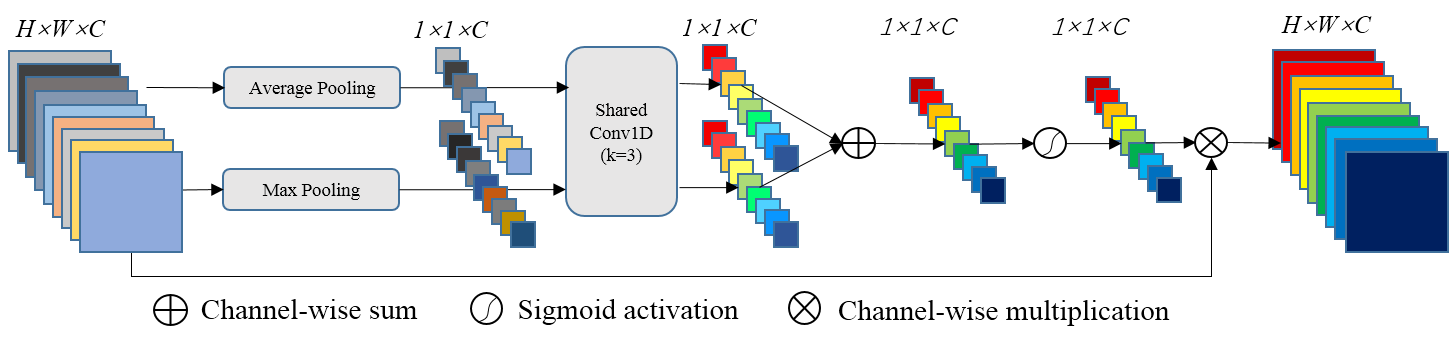}
    \caption{Top: the modified U-net architecture, where the skip connections are replaced by efficient channel attention (ECA) blocks. Bottom: details of the ECA block.}
    \label{fig:unet}
\end{figure}

Segmentation is a difficult task, as local information (sharp object boundaries) has to be combined with global information (overall shape of an object) to obtain a correct and precise mask. The sketch of the architecture is shown of Fig.~\ref{fig:unet}. The full sized input images are first rescaled to $192\times 256$, which is the input resolution of the CNN. It is then subjected to a cascading sequence of convolusions and maxpool-based coarsening, which enable local information to propagate to the global (only $6\times 8$ pixel) representation. This is followed by a similar sequence of convolutions with upscaling, which propagates global information gradually towards local, higher resolution representation. The original U-net architecture combines local (object boundary) with global (object recognition) information by employing skip connections at every internal resolution level. The performance of the network increases when the skip conncetions are replaced by an improved channel attention block called ``efficient channel attention'' (ECA) \cite{wang2020ecanet}.

\subsection{Feature classifiers}
\label{sec:feature-classifiers}

We designed implicit prefiltering methods to emphasize relevant features which expert dermatologists found decisive, like the criteria in the ABCD rule, or Menzies's list \cite{menzies1996frequency}. The image-based classifiers take as input either the original image, its subregion, or an image preprocessed in a specialized way. The preprocessed input is rescaled to a standard size ($256\times 256$ pixels), classified by a deep CNN (we choose DenseNet-201 for its performance\cite{huang2017densely}, see Section \ref{sec:results-backbone} for comparison of other backbones), optimized for categorical crossentropy loss with the Adam optimizer at learning rate $10^{-5}$, and trained with early stopping until the best validation accuracy is reached with patience of 10 epochs. 

The \emph{whole} classifier takes the original image. In principle this contains all known image-based information about the lesion, although fine details might be lost during rescale. Nevertheless, as we will see is Section \ref{sec:results-feature}, this classifier has the best individual performance of the five feature classifiers, and we take it as a baseline when assessing the potential benefits of our strategy of fusing multiple feature classifiers.

The \emph{border} classifier uses a thin stripe along the perimeter of the mask, which is 5 pixel wide in the outwards direction, and 20 pixel wide inwards. This is expected to be the closest match to the ``B'' (Border) criterion of the ABCDE rule.

The \emph{color asymmetry} classifier is designed to capture the asymmetry (criterion ``A'') in color and structure of the lesion. The input image to this classifier is prepared in the following way. First, the center of mass of the segmentation mask and its major and minor axes (the orthogonal eigenvectors of its 2D inertia tensor) are calculated. Then for a given axis the absolute value of the rgb color difference between the original image and its reflection over the axis is recorded.  The final input image is the average of three such differences: one over the mask major axis, one over the minor axis, and one over reflection over both axes. In this process the pixels outside the lesion are replaced by the average color of the lesion perimeter, so only the asymmetry of the color structure of the lesion gives contribution.

The fine details of the lesion, which are lost in the limited resolution of the whole image detector, are captured by the \emph{central} classifier. Simply a $256\times 256$ pixel region centered around the center of mass of the segmentation mask is taken at the original resolution of the lesion image. This is expected to capture fine details of the lesion, including brown or gray dots and globules and atypical pigment network.

The above four classifiers are trained individually for the final target: melanoma vs.\ not melanoma, based on preprocessed images with certain aspects highlighted. The fifth classifier is different: it is trained for the presence or absence of a specific diagnostic criterion: the \emph{blue-white veil}. This criterion is considered by dermatologists as one of the most useful in diagnosing melanoma: while only 51\% of melanoma samples carry this property, its presence is indicative at 97\% specificity \cite{menzies1996frequency}. The classifier was trained on the derm7pt dataset\cite{kawahara2019derm7pt}, where ground truth labels for the presence of various features including blue white veil are provided for each sample. The best performance (measured on a held out test set) was achieved when the preprocessed input image contained only the lesion (surroundings were masked out), and the images were augmented by adding a superpixel-based locally averaged\cite{wei2018boruvka} copy at superpixel count $10\,000$.

\begin{figure}
    \centering
    \includegraphics[width=0.65\textwidth]{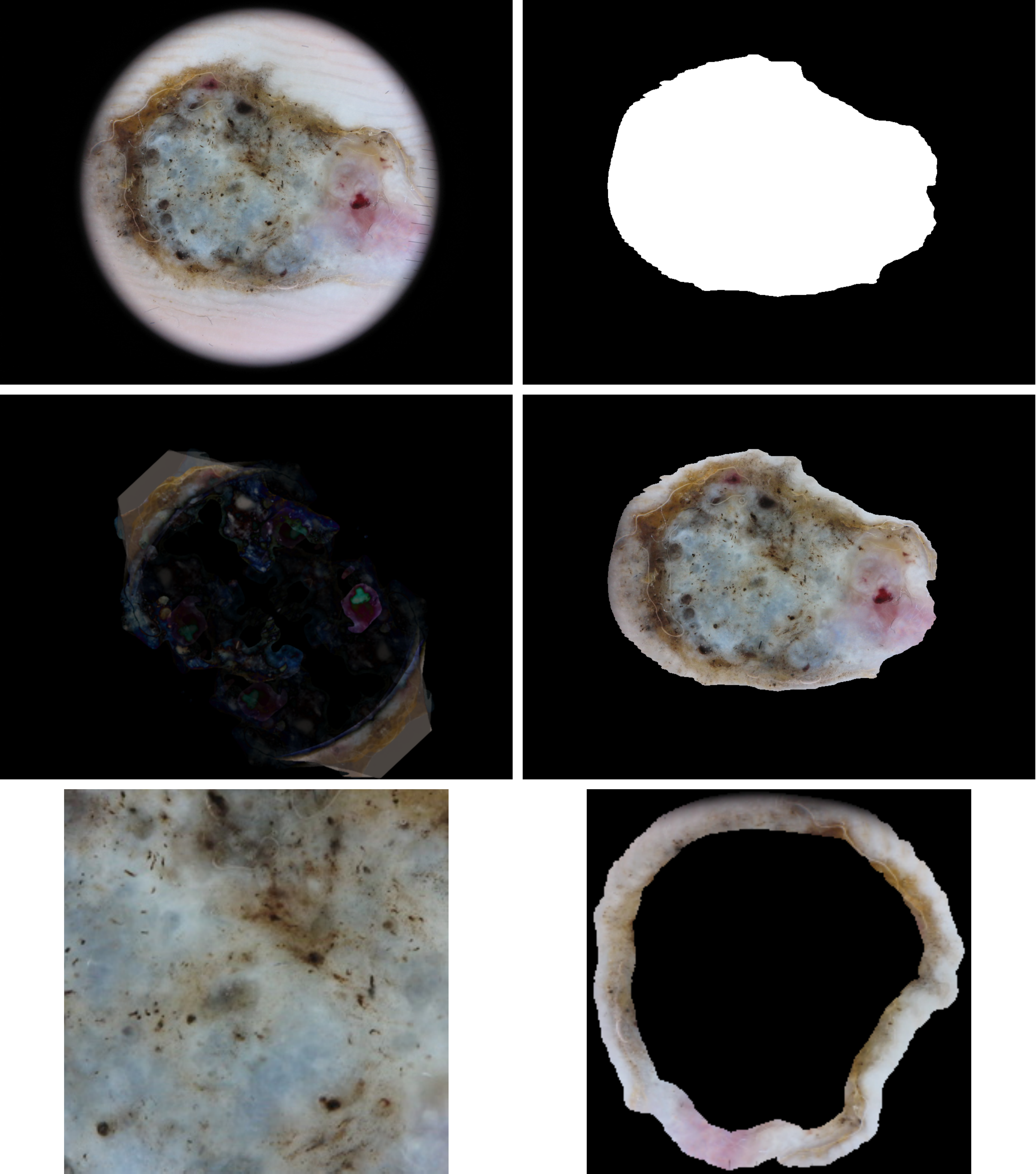}
    \caption{Preprocessed input images for the classifiers. The images from left to right, then top to bottom: original image (whole); binary mask used for the following images; input to color asymmetry classifier; masked image for blue white veil detector; center image; border image.}
    \label{fig:inputs}
\end{figure}

Samples from the preprocessed input images of the different classifiers are shown on Fig.~\ref{fig:inputs}.

The convnet performing the classification consists of a deep backbone (see Sec.~\ref{sec:results-backbone} for comparison), and a classifier head. The latter starts with global average pooling to contract spatial dimensions, followed by two hidden dense relu-activated layers of 4096 neurons each with dropout, and terminated by a two-unit (one for each class) softmax-activated layer yielding the soft decisions.

\subsection{Committee machine}
\label{sec:committee-machine}

The outputs of the feature classifiers are combined by a fully connected shallow neural net committee machine. We found the following architecture optimal: three hidden layers consisting of 128, 64 and 32 neurons, with dropout of 0.1 after each layer. During training, the categorical crossentropy loss is optimized by Adams optimizer with learning rate $10^{-5}$. Due to the small size of the net, the training is very quick (of the order of a minute on a competitive CPU); early stopping is used with patience 10 epochs.

To optimize the performance of the committee machine, we compared two kind of outputs of the feature classifiers: either used the final soft decision (a single softmax value per sample between 0 and 1), or the output of the previous, dense feature layer (4096 relu-activated values per sample).

We also considered using biased crossentropy loss during training, where false negative and false positive samples receive different penalty:
\begin{equation}
    \text{loss} = -\left[ b \cdot y \log(p) + (1-y) \log(1-p) \right] \,,
\end{equation}
where $y$ is the ground truth (0 for non-melanoma and 1 for melanoma), $p$ is the soft prediction, and when the bias $b$ is larger than 1, the false negative samples contribute to the loss with higher weight than the false positives.

\section{Results and discussion}\label{sec:res-disc}

To assess the performance of the components and the whole of our method, we performed a series of tests using the ground truth annotated training set released for the ISIC 2019 skin lesion classification challenge\cite{tschandl_ham10000_2018, codella_skin_2017, combalia_bcn20000:_2019}. We set the task as binary classification separating melanoma cases from all other lesions. As dataset is unbalanced containing only 18\% melanoma, care was taken to reduce its impact by employing class-weighting during model training and careful selection of the decision threshold.

\subsection{Comparison of segmenters}
\label{results-segmenter}

To assess the performance and reliability of the segmenter, we compared U-net without and with the ECA block, and trained on two different skin lesion datasets; the results are shown on Table~\ref{tab:segmenter}. The two datasets are PH2 \cite{mendonca2013ph2}, containing 200 dermoscopic images of common nevi, atypical nevi and melanoma, together with a pixel level mask of the lesions; and the training dataset for Task 1 of the ISIC2018 challenge \cite{codella_skin_2017,tschandl_ham10000_2018}, consisting of 2594 dermoscopic images of a variety of skin lesion types, together with the mask of the lesion.  In all cases we used the soft (differentiable) Jaccard distance as loss, Adam optimizer with learning rate $3\times 10^{-3}$, and trained either for 100 epochs (with the PH2 dataset) or used early stopping watching validation loss (on 10\% held-out part of the training set) with patience 10 epochs. At inference the soft (sigmoid) output coming out of the convnet at the $192\times 256$ pixel resolution is first rescaled to the resolution of the input lesion image, then thresholded at 0.5, resulting in a binary mask.

\begin{table}[h]
    \centering
    \begin{tabular}{ll|cc|cc|c}
         &  & \multicolumn{2}{c|}{PH2-25\%}  
            & \multicolumn{2}{c}{ISIC2018-20\%} & ISIC2019 \\
        Convnet & Train dataset & IoU & Dice & IoU & Dice & empty rate \\
        \hline
        U-net & PH2-75\% & \textit{0.853} & \textit{0.916} & 0.589 & 0.695 & 0.0058 \\
        U-net+ECA & PH2-75\% & \textit{0.851} & \textit{0.915} & 0.600 & 0.701 & 0.0019 \\
        U-net+ECA & ISIC2018-80\% & 0.822 & 0.897 & \textbf{0.812} & \textbf{0.883} & \textbf{0.0000}
    \end{tabular}
    \medskip
    \caption{
    Benchmark comparison of different versions of the architecture and training of the segmenter. While the models trained on PH2 perform slightly better on the PH2 held out test set, the performance on the $10\times$ larger ISIC2018 held out test set is much more relevant, as well as the reduction of the empty masks to almost zero on the ISIC2019 dataset we use for classification experiments in the next sections.
    }
    \label{tab:segmenter}
\end{table}

The original U-net architecture trained on 75\% of the PH2 dataset achieves good benchmark, 85\% intersection over union (IoU, i.e., intersection over union, a.k.a Jaccard index) on the remaining 25\% of the dataset. However, when it is tested against lesions from the ISIC2018 dataset, the results are poorer (only 59\% IoU). Also, when applied on the ISIC2019 dataset, which we are going to use for segmentation in the remainder of this paper, over 0.5\% of the cases the prediction was an empty mask, seriously impacting the further processing of the involved lesions.  After replacing the simple skip connections of U-net with ECA, the results generally improved: there was an $\approx 1\%$ increase in the IoU for the ISIC2018 dataset, and ever more importantly the number of empty masks for the ISIC2019 dataset reduced by a factor of 3. There was practically no change for the held-out test set of PH2, but since that contained only 50 images, no strong conclusions can be drawn.

Overall the best benchmark is achieved by the U-net+ECA architecture, when trained on 80\% of the larger ISIC2018 dataset, under heavy augmentation. The IoU for the held-out ISIC2018 test set jumped from 60\% to 81\%, and the number of empty masks for ISIC2019 lesions dropped to practically zero. We use this version of the segmenter in the rest of this paper.  It is interesting to note, that a slight drop has been observed when tested against the PH2 dataset (from 85\% IoU to 82\%), which we explain by database dependence. The grand truth masks of the PH2 dataset, drawn by expert dermatologists, are typically tighter than those of the ISIC2018 Task1 dataset, drawn almost certainly by different experts. This systematic trend is learned by the model, resulting in a slight decrease for the PH2 IoU despite the $10\times$ larger training set.

\subsection{Comparison of classifier backbones}
\label{sec:results-backbone}

A number of common deep convnet backbones have been compared. The goal here was to test raw image classification performance on skin lesions, therefore we used the original (whole) images as input.
In the first round of comparisions 9 models were trained with the unbalanced input dataset (no class weighting); the results are shown in Table~\ref{tab:backbones}. The DenseNet and ResNet families, especially the largest models, as well as the Xception model performed well.

\begin{table}[htb]
    \centering
        \begin{tabular}{lcccc}
    
classifier backbone & accuracy & balanced accuracy & false negative rate & ROC AUC \\
\hline
DenseNet121 & $0.8944\pm 0.0030$ & $0.7817\pm 0.0111$ & $0.3938\pm 0.0255$ & $0.9128\pm 0.0074$ \\
DenseNet169 & $0.8964\pm 0.0069$ & $0.7985\pm 0.0164$ & $0.3540\pm 0.0442$ & $0.9240\pm 0.0113$ \\
DenseNet201 & $0.9001\pm 0.0027$ & $0.8073\pm 0.0103$ & $0.3370\pm 0.0232$ & $0.9270\pm 0.0079$ \\
InceptionResNetV2 & $0.8998\pm 0.0107$ & $0.7919\pm 0.0261$ & $0.3761\pm 0.0523$ & $0.8922\pm 0.0468$ \\
InceptionV3 & $0.8972\pm 0.0046$ & $0.7978\pm 0.0132$ & $0.3569\pm 0.0320$ & $0.9151\pm 0.0061$ \\
ResNet50 & $0.9092\pm 0.0086$ & $0.8198\pm 0.0136$ & $0.3193\pm 0.0223$ & $0.9340\pm 0.0063$ \\
ResNet101 & $0.9085\pm 0.0074$ & $0.8119\pm 0.0086$ & $0.3385\pm 0.0133$ & $0.9367\pm 0.0081$ \\
ResNet152 & $0.9127\pm 0.0042$ & $0.8254\pm 0.0163$ & $0.3105\pm 0.0380$ & $0.9402\pm 0.0076$ \\
Xception & $0.9011\pm 0.0093$ & $0.8091\pm 0.0233$ & $0.3341\pm 0.0485$ & $0.9280\pm 0.0085$ \\

    \end{tabular}
    \medskip
    \caption{First round of comparison of classifier backbones on the original (whole) image. The displayed values and uncertainties are the average and standard deviation obtained from three independent trainings (starting from different set of random weights) for an 80\%-10\%-10\% training-validation-test data split. The accuracy, balanced accuracy and false negative rate are taken at the 0.5 threshold level.}
    \label{tab:backbones}
\end{table}

In the second round of comparisons the top three contenders were tested more thoroughly: using a 5-fold cross validation test of the skin dataset, and trained with class-based weights to reduce the effects of class imbalance. The results are displayed in Table~\ref{tab:backbones2}. In this comparison ResNet152 turned out to be the weakest, and DenseNet201 had the best overall performance, especially when considering ROC (Receiver Operating Characteristics) and AUC (Area Under The Curve). We used this backbone in the remainder of this paper.

\begin{table}[h]
    \centering
    \begin{tabular}{lcccc}
classifier backbone & balanced accuracy & false negative rate & ROC AUC & training time [hour:min] \\
\hline
DenseNet201 & $0.8453\pm 0.0114$ & $0.2388\pm 0.0167$ & $0.9335\pm 0.0111$ & 07:43 $\pm$ 03:17 \\
ResNet152   & $0.8288\pm 0.0158$ & $0.2758\pm 0.0223$ & $0.9260\pm 0.0122$ & 07:16 $\pm$ 02:41 \\
Xception    & $0.8455\pm 0.0076$ & $0.2377\pm 0.0159$ & $0.9298\pm 0.0051$ & 12:18 $\pm$ 03:05 \\
    \end{tabular}
    \medskip
    \caption{Second round of comparison of classifier backbones on the whole images. The displayed values and uncertainties are the average and standard deviation obtained by a 5-fold cross validation test. The balanced accuracy and false negative rate are taken at the 0.5 threshold level. The training time is indicated for an Nvidia GeForce GTX 1080 GPU, for a given backbone the spread is caused by the early stopping invoked after different number of epochs in the cross validation folds.}
    \label{tab:backbones2}
\end{table}

\subsection{Feature classifiers}
\label{sec:results-feature}

Table~\ref{tab:results-feature} shows the benchmark results of the five feature classifiers. To reduce the impact of the unbalanced dataset, during training we used class-based weighting. As expected, the classifier processing the whole image is the best in all metrics, but the color asymmetry, center and border based classifiers are also providing good results. 

\begin{table}[h]
    \centering
    \begin{tabular}{lcccc}
    
feature classifier & accuracy & balanced accuracy & false negative rate & ROC AUC \\
\hline
whole & $0.8995 \pm 0.0189$ & $0.8453 \pm 0.0114$ & $0.2388 \pm 0.0167$ & $0.9335 \pm 0.0099$\\
color asymmetry & $0.7982 \pm 0.0266$ & $0.7546 \pm 0.0237$ & $0.3131 \pm 0.0932$ & $0.8435 \pm 0.0162$\\
center & $0.8751 \pm 0.0132$ & $0.7901 \pm 0.0123$ & $0.3421 \pm 0.0369$ & $0.8988 \pm 0.0057$\\
border & $0.8297 \pm 0.0209$ & $0.7532 \pm 0.0136$ & $0.3658 \pm 0.0574$ & $0.8553 \pm 0.0096$\\
blue white veil & $0.7975 \pm 0.0050$ & $0.6069 \pm 0.0098$ & $0.6895 \pm 0.0182$ & $0.6474 \pm 0.0110$\\

    \end{tabular}
    \medskip
    \caption{Benchmark results for the individual feature classifiers. The displayed values and the uncertainties are the average and the standard deviation obtained by a 5-fold cross validation test.
    The accuracy, balanced accuracy and false negative rate are taken at the 0.5 threshold level.\\
    $^*$The blue white veil traits detector, unlike the others, is trained for the target of blue white veil presence, and evaluated here for melanoma.}
    \label{tab:results-feature}
\end{table}

The blue white veil traits detector, unlike the other four, is trained on the derm7pt dataset\cite{kawahara2019derm7pt} for the target of blue white veil presence. In Table~\ref{tab:results-feature} the metrics are evaluated for non-melanoma / melanoma classification, and since only about half of the melanoma samples carry this mark, this detector is not expected to perform as good as the others. However, its output is still useful when combined together in the committee machine.

\subsection{Fusion by committee machine}
\label{sec:results-committee}

In the proposed model the results of the individual feature classifiers are combined by a shallow net committee machine. Two options are compared: (i) the final soft decision value (``softmax'') of the feature classifiers are used as committee machine input, and (ii) the output of the previous relu-activated dense feature layer (``feature layer'') of the feature classifiers are used. Table~\ref{tab:committee} shows the benchmark results: while the committee machine with the softmax input is just barely better than the best individual feature classifier (the ``whole'' classifier), when the feature layer input is used, the results show a definite improvement. In all cases class-weighted training was employed, and the decisions were taken at the 0.5 threshold level.

\begin{table}[h]
    \centering
    \begin{tabular}{lcccc}
    
com. mach. input&accuracy&balanced accuracy&false negative rate&ROC AUC\\
\hline
softmax&$0.9161\pm 0.0098$&$0.8500\pm 0.0220$&$0.2529\pm 0.0545$&$0.9408\pm 0.0029$\\
feature layer&$0.9231\pm 0.0079$&$0.8448\pm 0.0090$&$0.2770\pm 0.0197$&$0.9475\pm 0.0058$\\

    \end{tabular}
    \medskip
    \caption{Benchmark results obtained by the committee machine at the 0.5 threshold level for different input data. The displayed values and uncertainties are the average and standard deviation obtained by a 5-fold cross validation test. To further improve statistics (reduce the error in the averages), for each fold 5 independent training of the committee machine took place starting from a different set of random weights.}
    \label{tab:committee}
\end{table}

\subsection{ROC curve analysis and controlling false negative rate}
\label{sec:results-roc}

To have a better control on the selection of the decision threshold and the reduction of the false negative rate, it is useful to consider the ROC curve, which plots the true positive rate as a function of the false positive rate as the decision threshold is scanned between 0 and 1. AUC provides a global indication of the performance of the classifier, where all threshold levels are taken into account. It also enables to select an optimal decision threshold instead of the naive 0.5 level based on different requirements, like optimizing the balance accuracy (derivative of the curve is 1), or operating at a fixed false negative rate (equivalently at fixed true positive rate).

\begin{figure}[ht]
    \centering
    \includegraphics[width=0.5\textwidth]{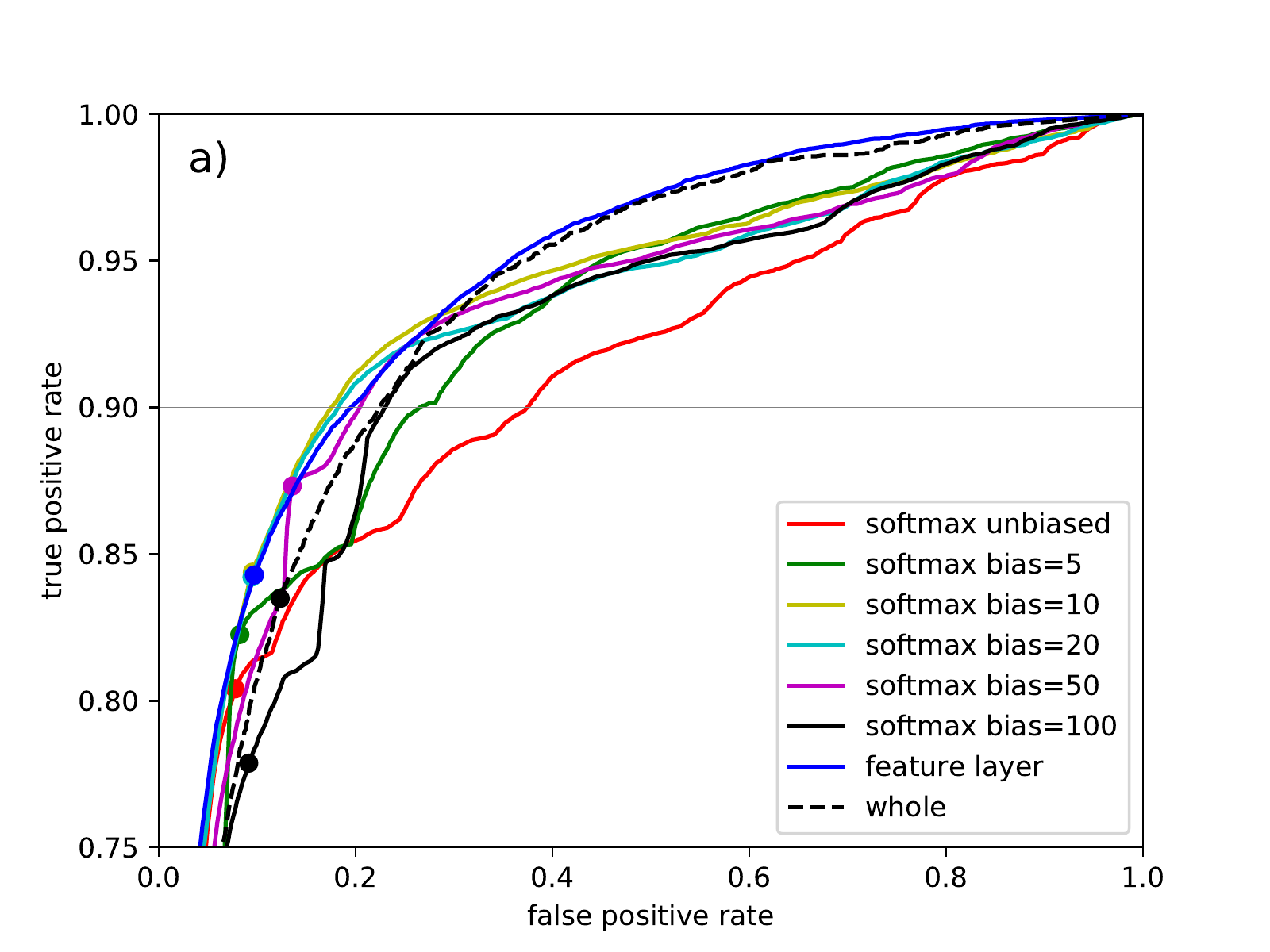}%
    \includegraphics[width=0.5\textwidth]{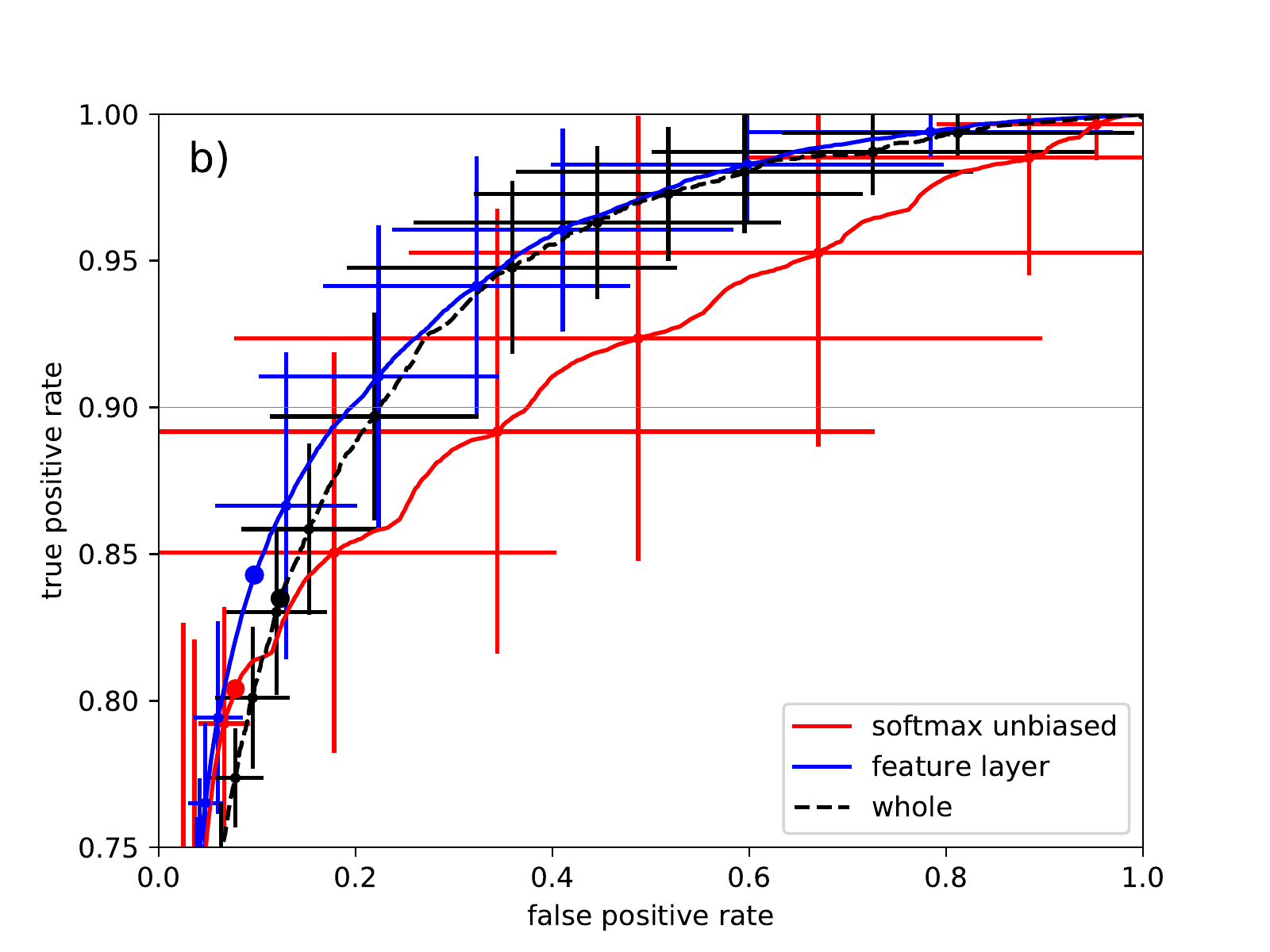}
    \caption{(a) The top quarter of the ROC curve for different input and training of the commitee machine. On each curve the point corresponding to the best balanced accuracy is marked by a filled circle, and the 10\% false negative rate is shown by the thin gray line. (b) The variation of the false and true positive rates for a given threshold value, displayed as population standard deviation at selected points.}
    \label{fig:roc}
\end{figure}

\begin{table}[h]
    \centering
    \begin{tabular}{lcccc}

committee machine&ROC AUC&best balanced accuracy&fpr at fnr=0.1\\
\hline
softmax unbiased&$0.9408\pm 0.0029$&$0.8632$&$0.3743$\\
softmax bias=5&$0.9413\pm 0.0054$&$0.8701$&$0.2653$\\
softmax bias=10&$0.9429\pm 0.0014$&$0.8743$&$0.1751$\\
softmax bias=20&$0.9426\pm 0.0016$&$0.8737$&$0.1823$\\
softmax bias=50&$0.9402\pm 0.0143$&$0.8688$&$0.2042$\\
softmax bias=100&$0.8857\pm 0.1988$&$0.8436$&$0.2301$\\
feature layer&$0.9475\pm 0.0058$&$0.8728$&$0.1950$\\
whole&$0.9335\pm 0.0111$&$0.8557$&$0.2248$\\

    \end{tabular}
    \medskip
    \caption{The ROC area under the curve, the best balanced accuracy, and the false positive rate obtained at 10\% false negative rate for different input and training of the committee machine. The best balanced accuracy and the false positive rate at the selected level are calculated from the averaged ROC curve, see Fig.~\ref{fig:roc}(a).}
    \label{tab:roc}
\end{table}

Figure~\ref{fig:roc}(a) displays the ROC curve for the committee machine with either softmax or feature layer input, and the whole classifier as baseline. In an attempt to reduce the false negative rate, results for the committee machine trained with biased loss is shown as well. The benchmark results are also tabulated in Table~\ref{tab:roc}.

A number of interesting conclusions can be drawn from the results. The training using biased loss improves the softmax-based committee machine's performance as the bias $b$ is increased up to around 10. ROC AUC, which is symmetric for false positive and false negative errors, also shows a very slight improvement, which is probably still the consequence of the original unbalanced dataset. When $b$ is increased to very high values like 100, serious degradation is observed. The baseline whole classifier, while having worse balanced accuracy, performs surprisingly well in the low false negative rate zone, beating the softmax-based committee machine. The overall winner, however, is the feature layer based committee machine. Training with biased loss did not affect the feature layer based committee machine, the results depend only negligibly on $b$.

The apparent discrepancy between the relative values of the ROC AUC for the unbiased softmax-based committee machine and the whole classifier (in the table the AUC for the softmax-based committee machine is slightly higher, while in the figure for most false positive rate values its curve is significantly lower than that of the whole classifier) can be resolved by considering how the averaging is applied. The displayed value and uncertainty in Table~\ref{tab:roc} is obtained by a 5-fold cross validation test, where for each fold 5 independent training of the committee machine took place with different set of random starting weights. For each 25 cases the area under the ROC curve was calculated and than averaged. In Fig.~\ref{fig:roc}(a) however, the curves are averaged first (averaging the false and true positive rates for a given threshold value), and then the area calculated afterwards. When the variation of the true and false positive rates for a given threshold value is large, averaging significantly lowers the curve, because it is concave. As can be seen in the Fig.~\ref{fig:roc}(b), for the softmax-based committee machine this variation is large, for the feature layer based committee machine it is small, and is intermediate for the baseline whole classifier. The small variability seen for the feature layer based committee machine is an extra indication for its robustness.

\subsection{Ablation study}
\label{sec:results-ablation}

To assess the contribution of the individual feature classifiers an ablation study was performed on the committee machine based on feature layer input, see Table~\ref{tab:ablation}. Compared to the full model, as expected the largest degradation is observed when the whole classifier is removed, as it contains most of the information. The second largest hit is observed with the removal of the center classifier, showing that the fine details of the image also carries useful information.  The removal of the border classifier causes very small performance loss, on the edge of detectability.  However, the remaining two detectors, highlighting the color asymmetry and the blue white veil traits, do not contribute individually to the performance of the committee machine. These two detectors would benefit from future refinement when considered as part of the committee machine.

\begin{table}[h]
    \centering
    \begin{tabular}{lccc}

committee machine&ROC AUC&best balanced accuracy&fpr at fnr=0.1\\
\hline
all &$0.9475\pm 0.0058$&$0.8728$&$0.1950$\\
all except whole&$0.9252\pm 0.0042$&$0.8456$&$0.2639$\\
all except color asymmetry&$0.9478\pm 0.0062$&$0.8743$&$0.1842$\\
all except center&$0.9413\pm 0.0057$&$0.8583$&$0.2392$\\
all except border&$0.9465\pm 0.0063$&$0.8715$&$0.1903$\\
all except blue white veil traits&$0.9474\pm 0.0057$&$0.8711$&$0.1974$\\

    \end{tabular}
    \caption{Performance of the feature layer based committee machine, when one feature classifier is removed from its input. }
    \label{tab:ablation}
\end{table}

While the contribution of the color asymmetry, border and blue white veil detectors are less significant, they are useful as indications of specific features characteristic of melanoma, as will be shown in the next Section.

\subsection{Indications by the feature classifiers}
\label{sec:results-indications}

The individual feature classifiers can be exploited to provide an indication for the presence or absence of the appropriate feature: whether the lesion has irregular border (by the border classifier), whether its color structure is asymmetric (color asymmetry), whether fine details are indicative or not for melanoma (center classifier), whether blue white veil is present (its detector), and finally whether the overall appearance of the lesion is characteristic or not for melanoma (the whole image based classifier).

We found that a useful quantification can be obtained by considering the soft prediction of the given classifier, and expressing it as a percentile on the distribution over a balanced dataset. To calculate the indications for a sample from the ISIC2019 dataset, we take the cross validation fold in which the given sample is in the test set, and evaluate the soft prediction for the sample using the classifier trained on that fold. Finally the validation set of the given fold is augmented by repeating the minority class enough times to get a balanced dataset between melanoma and non-melanoma, and express the sample's soft prediction as a percentile over the distribution of soft predictions for the augmented validation set.

\begin{figure}[htp]
    \centering
    \includegraphics[width=0.75\textwidth]{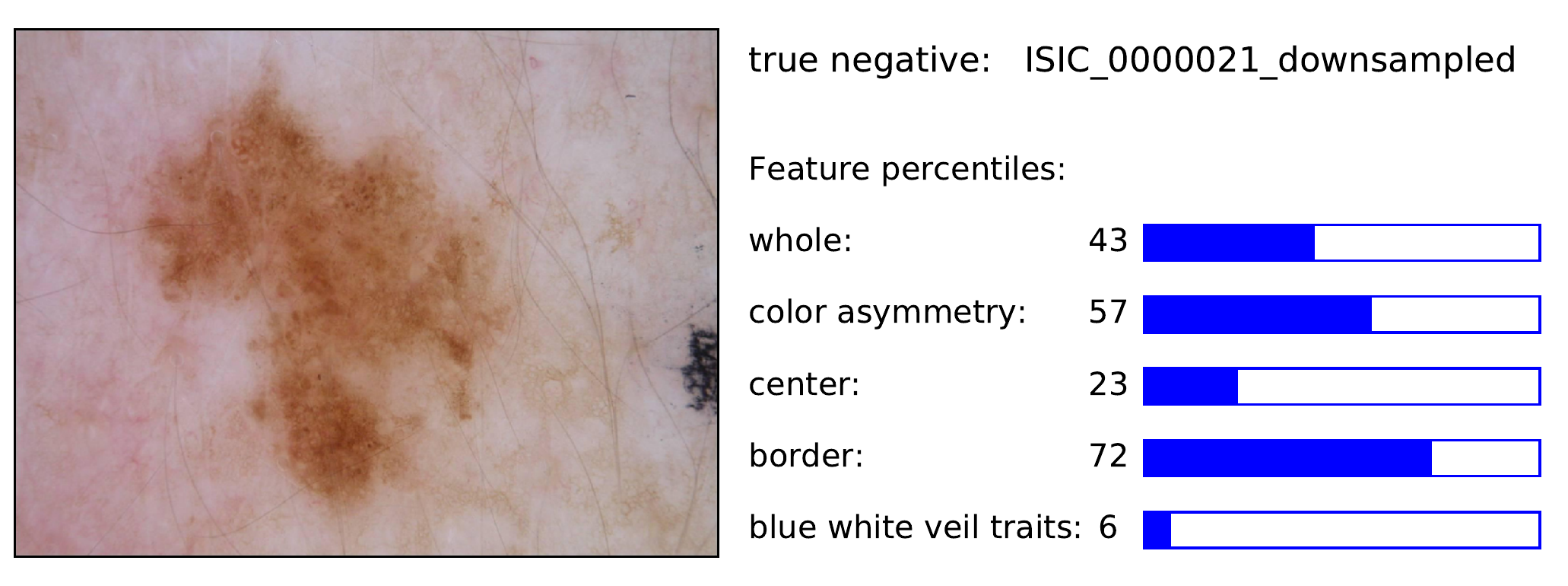}
    \includegraphics[width=0.75\textwidth]{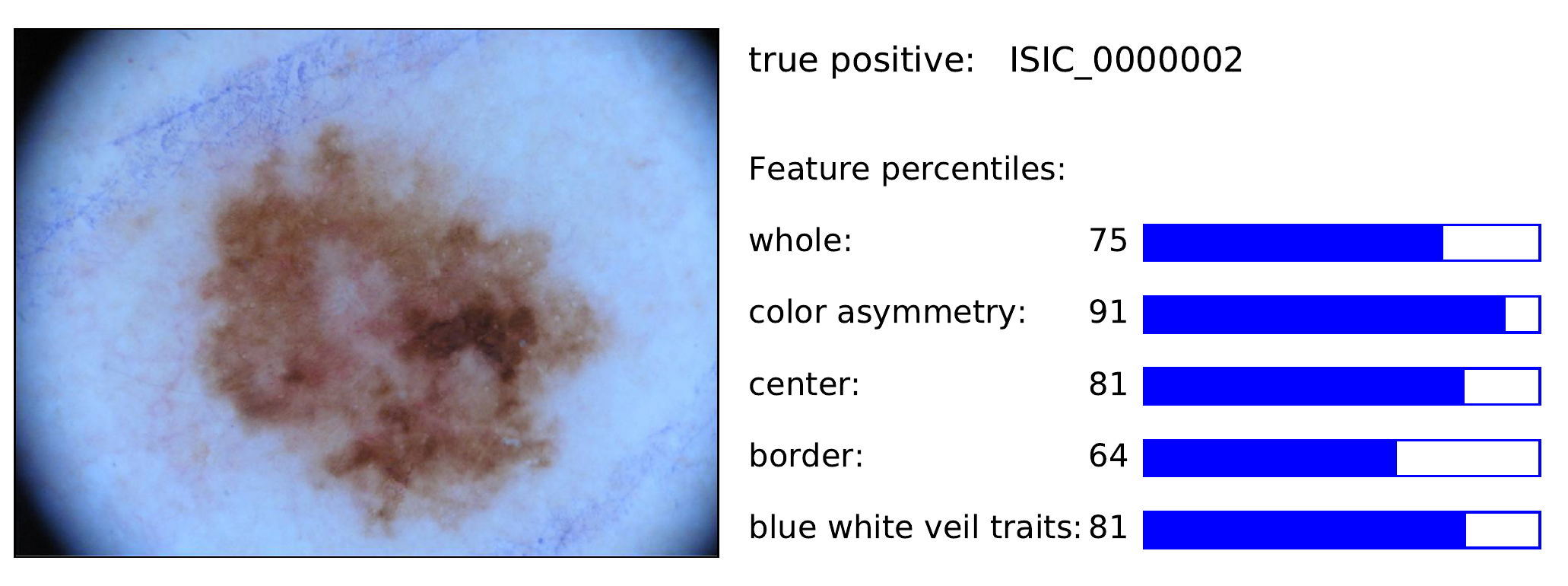}
    \includegraphics[width=0.75\textwidth]{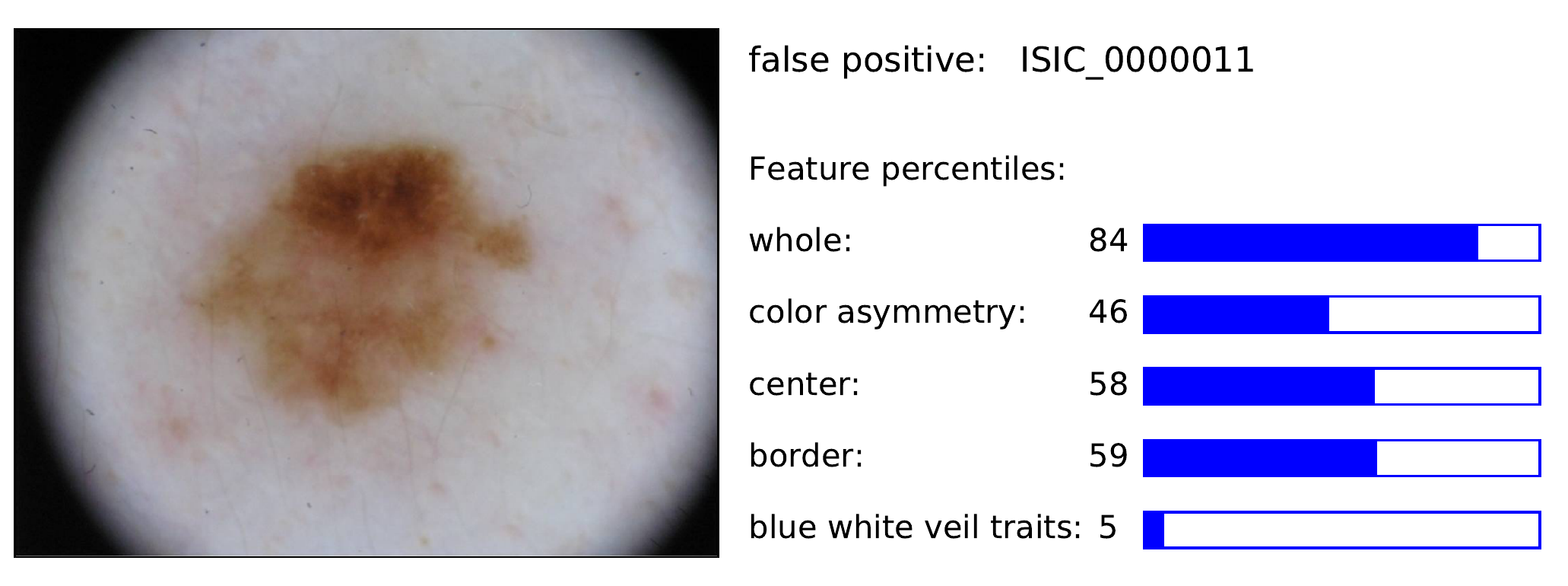}
    \includegraphics[width=0.75\textwidth]{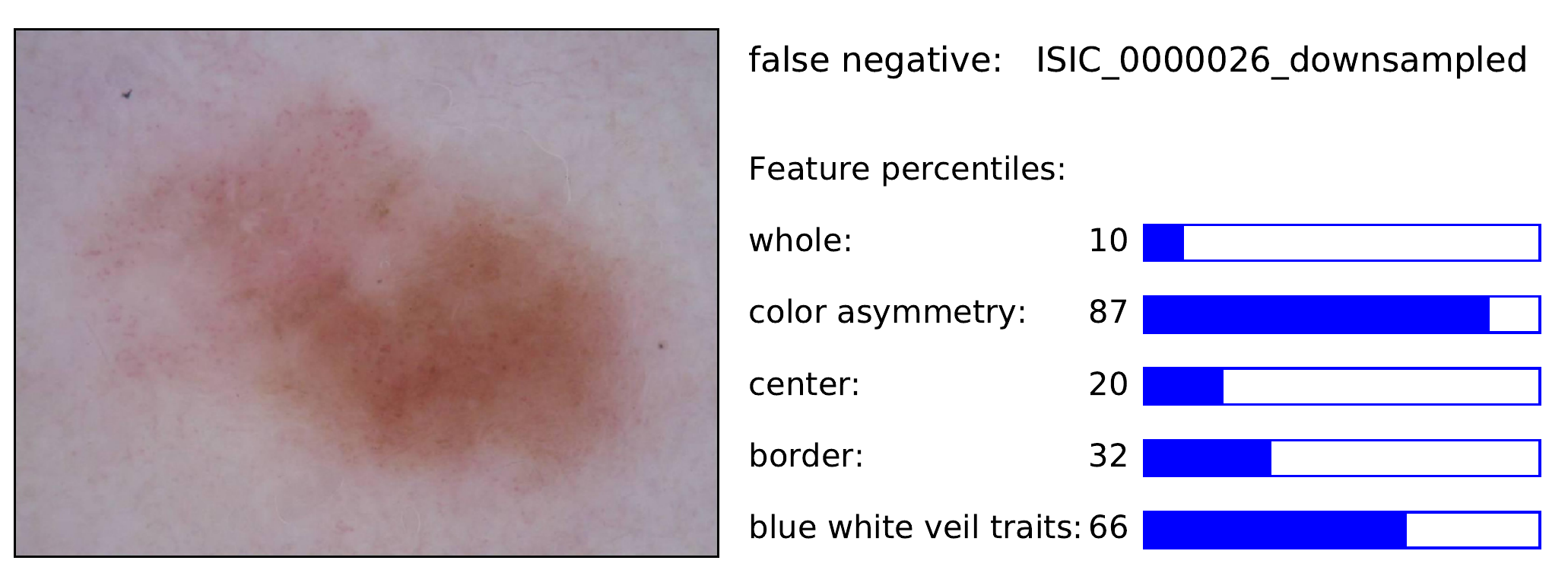}
    \caption{Indications of the feature classifiers quantified by the prediction percentiles.  Four examples are shown, labeled by the success or failure of the committee machine with feature layer input, with threshold set at 10\% false negative rate.}
    \label{fig:indications}
\end{figure}

We illustrate this method by two sets of examples. In the first set one sample each from the true negative, true positive, false positive and false negative cases are taken for the feature layer based committee machine thresholded at 10\% false negative rate. To prevent hand picking, the first sample in alphabetic order over its file name is taken for each category (e.g. false negative). The results are shown in Fig.~\ref{fig:indications}.

While not at the level of explainability (the percentiles do not determine the committee machine's classification outcome), they still can provide a useful indication on the selected aspects. For example in Fig.~\ref{fig:indications} the ragged borders of the true negative and true positive samples are indicated by high border figure, while the relatively smooth false negative example received lower border indication. The false negative is indeed a difficult case for the model: it is rather structureless (as can be seen from the low center percentile), and the only indication for melanoma is its highly asymmetric color structure (as indicated by the high color asymmetry percentile). The blue white veil traits detector indicates high value, 81 percentile, for the true positive case, while it is not obviously present on the lesion. However, true blue white veil is only expected in 50\% of melanoma cases, therefore only in the (top) 25\% of the balance dataset; so the 81 percentile value would be just barely indicative. But at the same time we cannot exclude that the blue white veil traits detector, trained on the relatively small derm7pt dataset, is picking up collateral features typically appearing together with blue white veil, but not always (as in the case of our true positive example).

The second set of examples consist of 25 melanoma samples, which were considered difficult for machine learning applications. Many of the samples are true positives for our feature layer based committee machine (at 10\% false negative rate), but some are false negatives.  To compare the prediction of our model and the indicative percentiles with observation by human experts, we performed a blind test in which two dermatology experts were asked to classify these lesions (without knowing that they are all melanoma), and to provide a list of features on which their decision is based. The results are shown in section 1 of the supplementary materials.

\subsection{Comparison with a weaker backbone}
\label{sec:results-densenet169}

For comparison we repeated the analysis of Sections \ref{sec:results-feature}-\ref{sec:results-ablation} with a weaker backbone: DenseNet169; the results are in section 2 of the supplementary materials. The conclusions are similar: the committee machine based on the softmax output of the feature classifiers outperforms the whole classifier, and the best performance is obtained by the committee machine using the feature layer outputs; all slightly lower (typically 1\% lower) than in the case of the DenseNet201 backbone.  Slight differences include that for the weaker backbone the ROC curve of the feature-layer based committee machine is uniformly better (at all thresholds) than the softmax based committee machine with any bias, while for the DenseNet201 backbone there are some parts of the curve (e.g., at 0.1 false negative rate) where the softmax based committee machine trained with bias 5 or 10 is slightly better.

\subsection{Deep networks and explanatory power for experts}

The strength of the deep learning becomes clear from the results of the whole image based classification that gives the best results. On the other hand, border irregularities, estimated presence of blue-white veil, level of color asymmetry, classification based on the center only all provide information for the human expert on understanding the underlying reasons of the classification as indicated by Fig.~\ref{fig:indications}. Dependencies between these factors are complex and non-linear estimations are in need that we did by means of a shallow network.

\section{Conclusions and outlook}\label{sec:conc-outlook}

In summary, we presented an ensemble-based method to classify skin lesions, which combines 5 feature classifiers by a learned committee machine. The 5 constituent classifiers were hand designed and implicit to enhance features considered important for melanoma by dermatologists. Our method shows improved performance over an image-based classifier. Out of the 5 feature classifiers, two (the whole image and the center image, with complementary strengths) are responsible for most of the improvements. We expect that enhancing feature engineering for those classifiers, as well as adding new ones will be able to overcome the lack of samples in the highly demanding skin lesion classification task, a potential key to further efficiency gains.

We also showed that indicative scores computed by the feature classifiers can provide useful insight into the various features on which the decision can be based.

We expect that the accuracy of the melanoma detection can be further improved, especially the dangerous false negative rate reduced, by both using more computational resources (employing larger, more potent convnet backbones, operating at higher image resolution), and also by further refining the feature engineering of the individual feature detectors.

Our last notes that justify our studies are as follows: (a) the dataset of ISIC 2019, the largest available public database, contains data mostly from white people, increasing the risk of estimation for other races and the individual features may help to reduce such risks. (b) It is known that additional side information about the history of the patient can lower misclassification rate and such information could be included into the training datasets. (c) The risk of misclassification of melanoma could be decreased by adding temporal information about the changes of the mole, i.e., exploiting condition E of the ABCDE scheme.  

\begin{acks}
E.S, K.F., A.U. and A.L. were supported by Application-specific highly reliable IT solutions project (Project no. ED-18-1-2019-0030) of the National Research, Development and Innovation Fund of Hungary under the Thematic Excellence Programme funding scheme. Additional funding was provided by the European Union co-financed by the European Social Fund (EFOP-3.6.3-16-2017-00002) and the Ministry of Innovation and Technology NRDI Office within the framework of the Artificial Intelligence National Laboratory Program.

\end{acks}

\bibliographystyle{ACM-Reference-Format}
\bibliography{skin}

\end{document}


\author{E Somfai, B Baffy, K Fenech, C Guo, R Hosszú, D Korózs, F Nunnari, M Pólik, D Sonntag, A Ulbert, A Lőrincz}
\date{2020}
\title{Supplementary materials \\
Minimizing false negative rate in melanoma detection and providing insight into the causes of classification}

\maketitle

\section{Blind test results}

We prepared and performed a blind test with two dermatology experts. We selected 25 lesion images from the ISIC 2019 dataset, and asked the experts independently to classify them as either melanoma (``MEL'') or non-melanoma (``non-MEL''), which is typically nevus; and provide a list of features on which their decision is based.  As our goal is the correct detection of melanoma, all 25 images were melanoma cases (which the experts did not know during the blind test).
\bigskip

The following table contains:
\begin{list}{\textbullet}{\parsep=0mm \topsep=1ex}
    \item the lesion images,
    \item the prediction of our committee machine (using feature layer input, and decision threshold set at 10\% false negative rate), and
    \item the classification by each expert together with their confidence level, and
    \item the list of lesion features on which the decision is based upon.
\end{list}

\def\tablecaption{Blind test results.}


\begin{table}[]
    \begin{tabular}{c c p{45mm} p{45mm}}
        Image & Ground truth and predictions & Expert 1 & Expert 2 \\
        \hline
        
\rule{0mm}{10mm}\raisebox{-28mm}{\includegraphics[height=30mm]{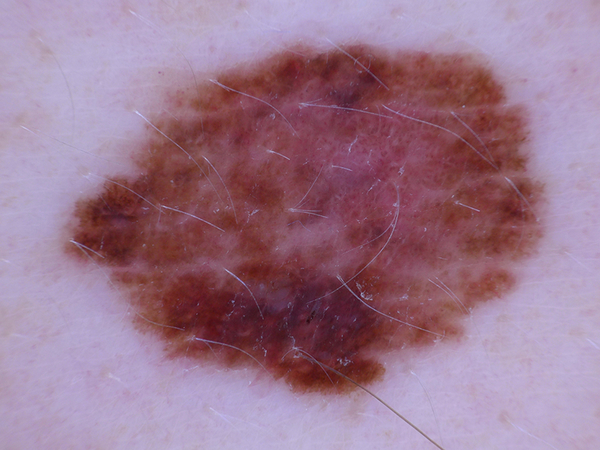}} & \raisebox{-26mm}{\includegraphics[height=28mm]{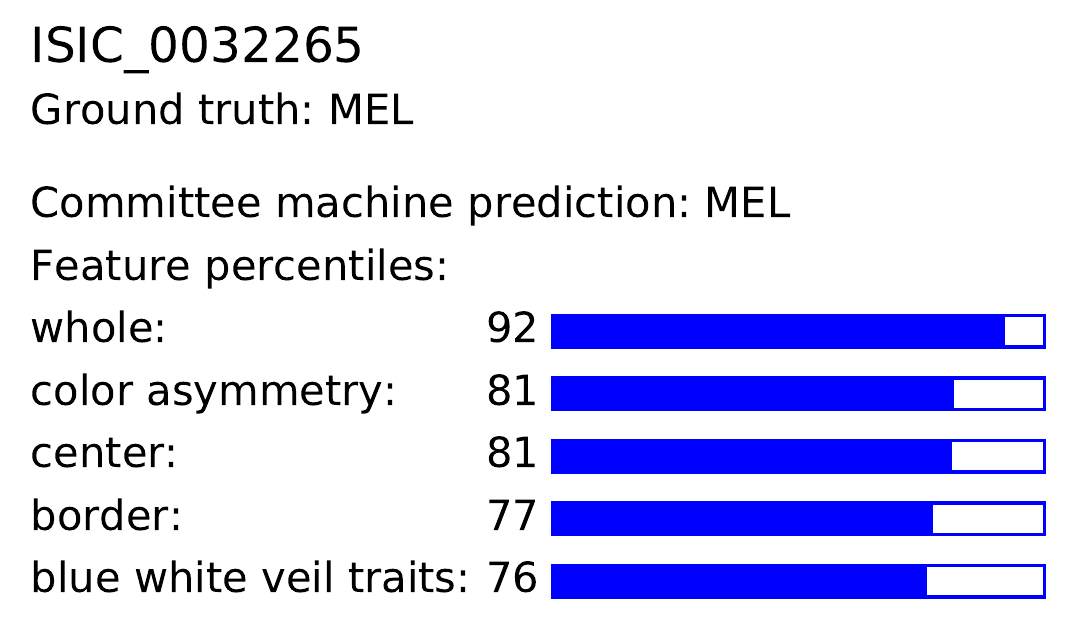}} & MEL (certain)\newline blue-white veil, peripheral black dots/globules, horizontal lines, scarline depigmentation, atypical capillaries & MEL (almost certain)\newline multiple blue-gray dots, 5-6 colors, peripheral black dots/globules, scar-like depigmentation, radial streaming, pseudopods, blue-white veil \\
\rule{0mm}{10mm}\raisebox{-28mm}{\includegraphics[height=30mm]{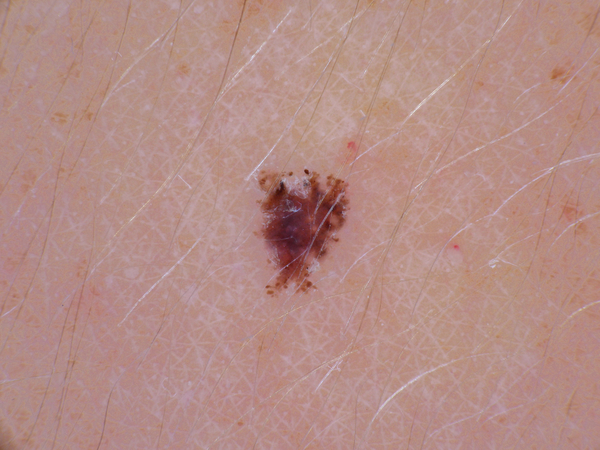}} & \raisebox{-26mm}{\includegraphics[height=28mm]{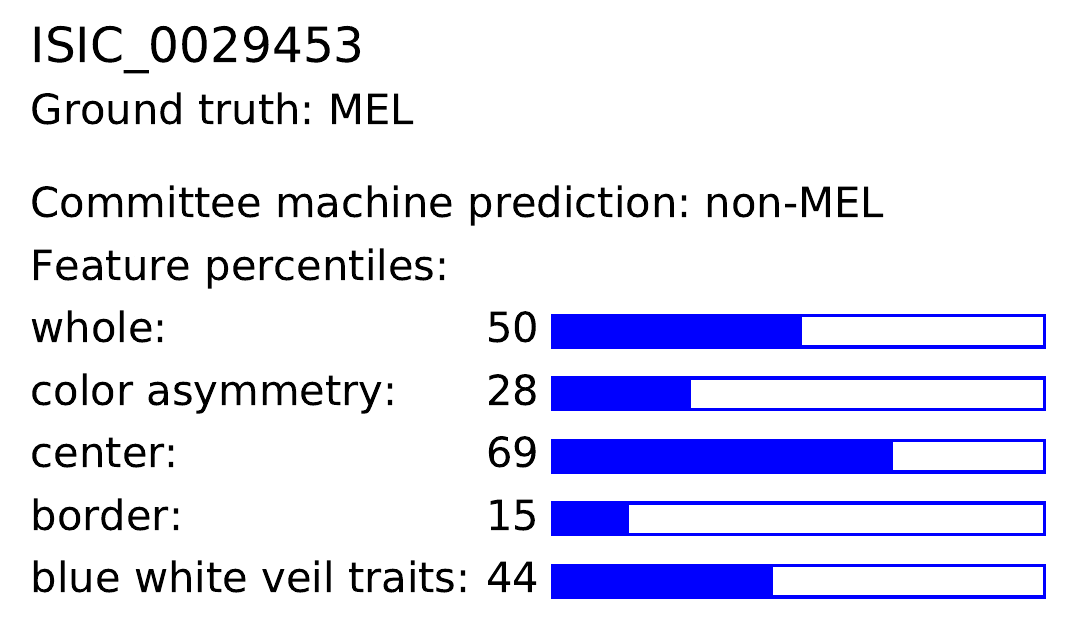}} & MEL (certain)\newline blue-white veil, visible big globules, pseudopods & MEL (certain)\newline peripheral black dots/globules, pseudopods, multiple brown dots, blue-white veil \\
\rule{0mm}{10mm}\raisebox{-28mm}{\includegraphics[height=30mm]{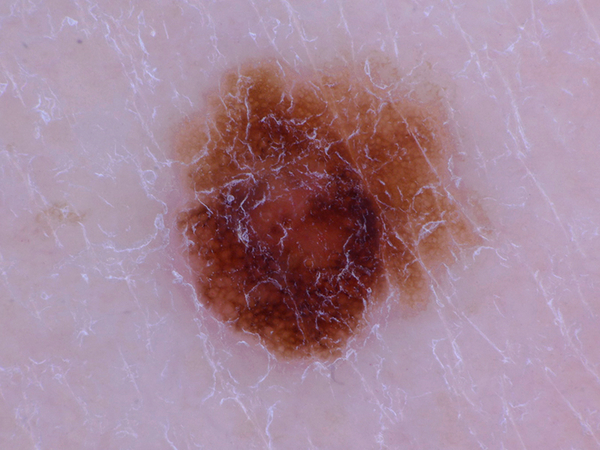}} & \raisebox{-26mm}{\includegraphics[height=28mm]{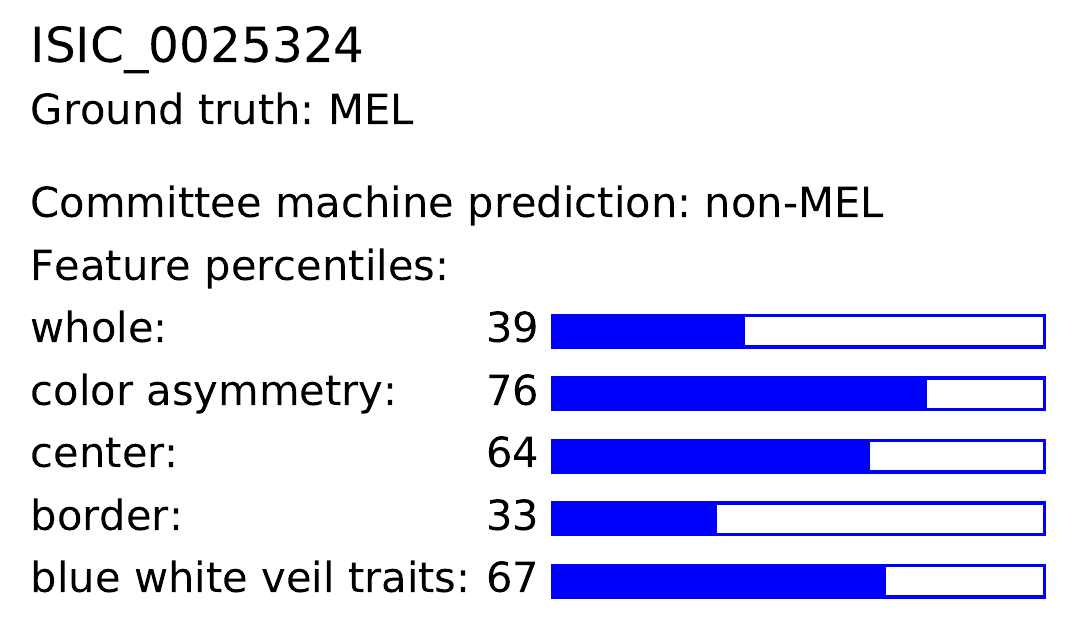}} & non-MEL (almost certain)\newline could be atypical nevus, small dots, but nothing else refers to melanoma & MEL (almost certain)\newline broadened network \\
\rule{0mm}{10mm}\raisebox{-28mm}{\includegraphics[height=30mm]{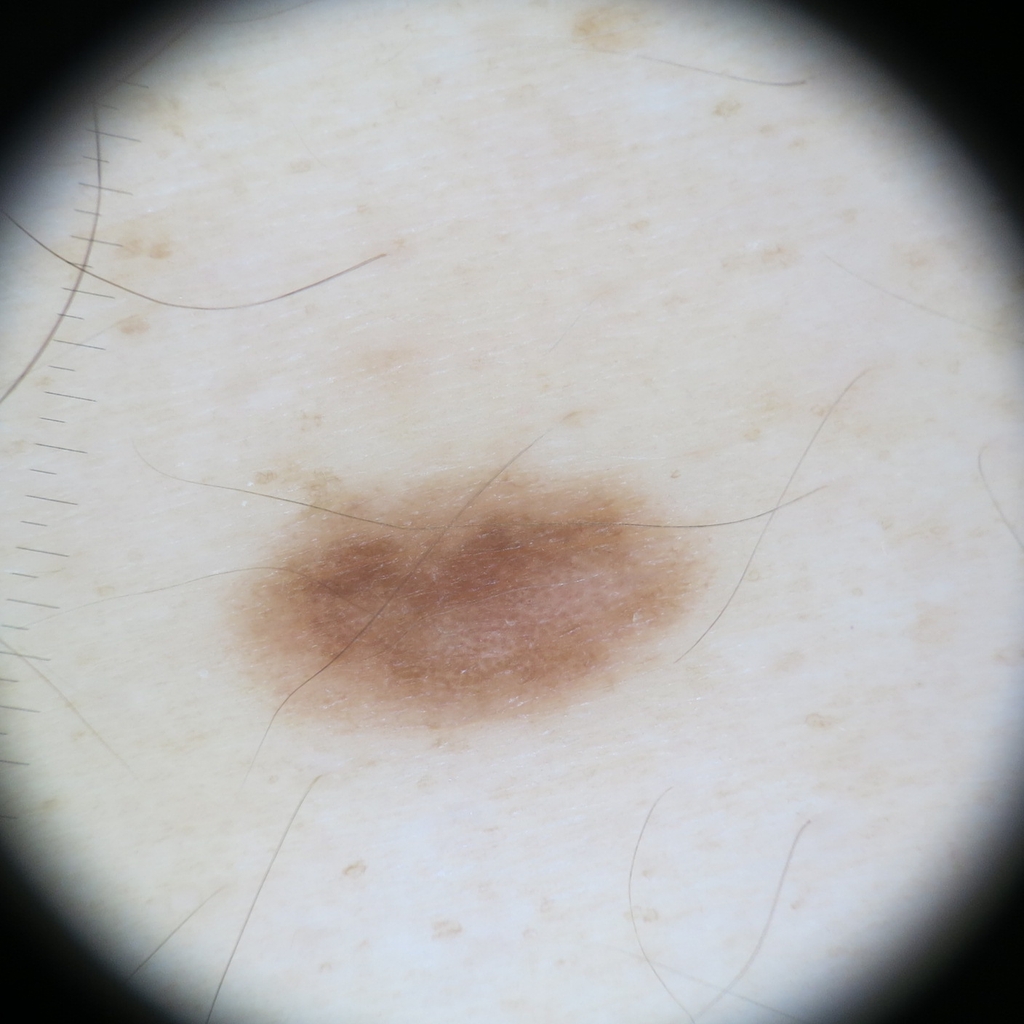}} & \raisebox{-26mm}{\includegraphics[height=28mm]{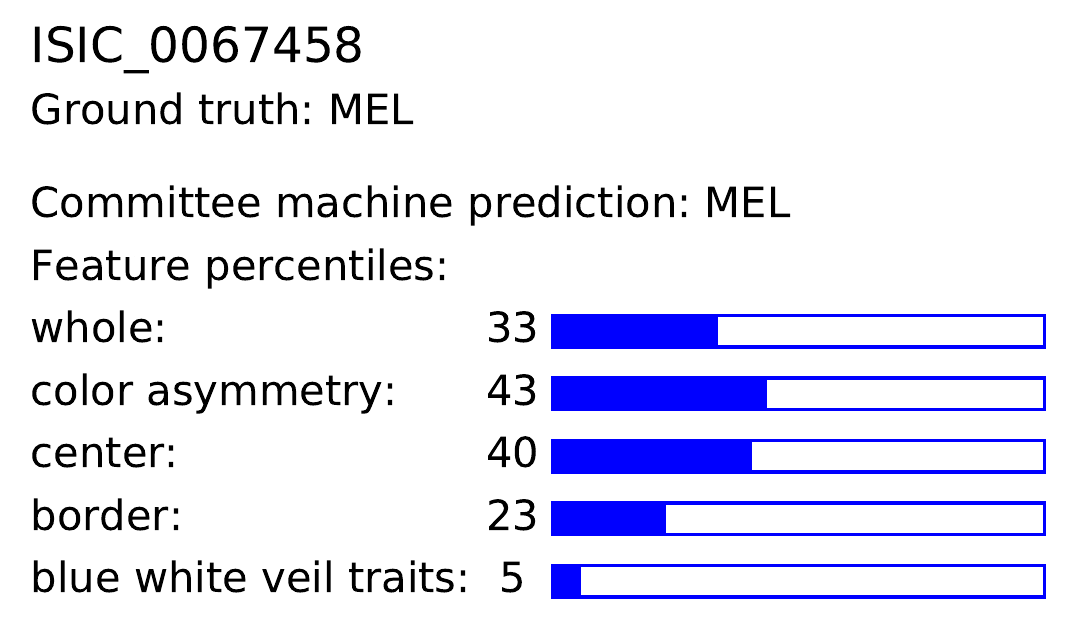}} & non-MEL (almost certain)\newline homogeneous color, in the middle maybe regressive part, it could be dysplastic nevus & non-MEL (certain)\newline single color, symmetric pigmentation pattern \\
\rule{0mm}{10mm}\raisebox{-28mm}{\includegraphics[height=30mm]{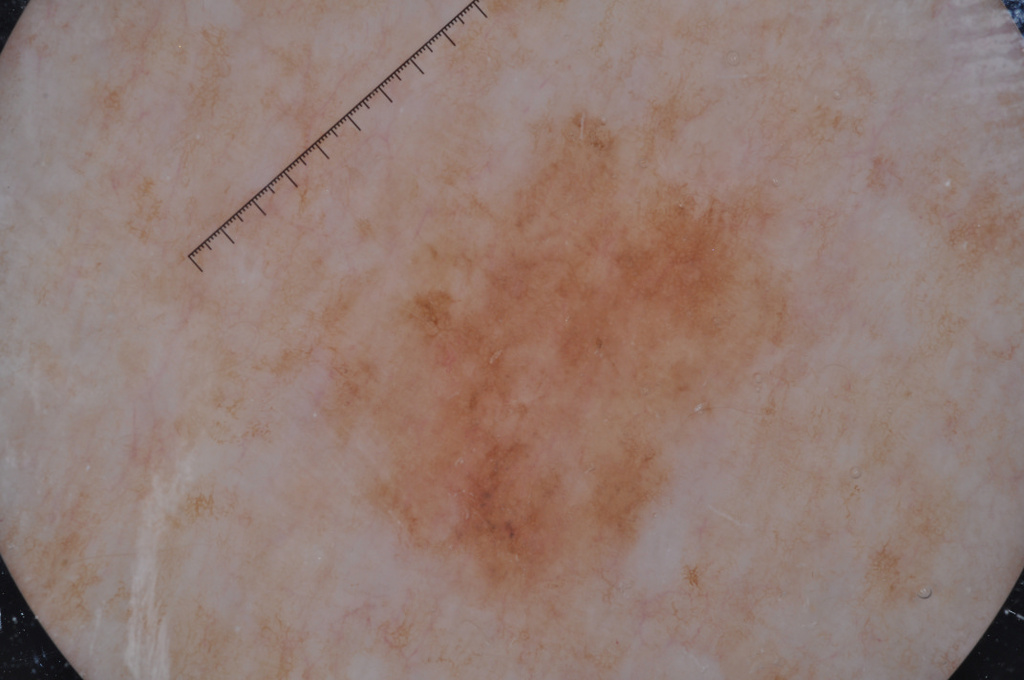}} & \raisebox{-26mm}{\includegraphics[height=28mm]{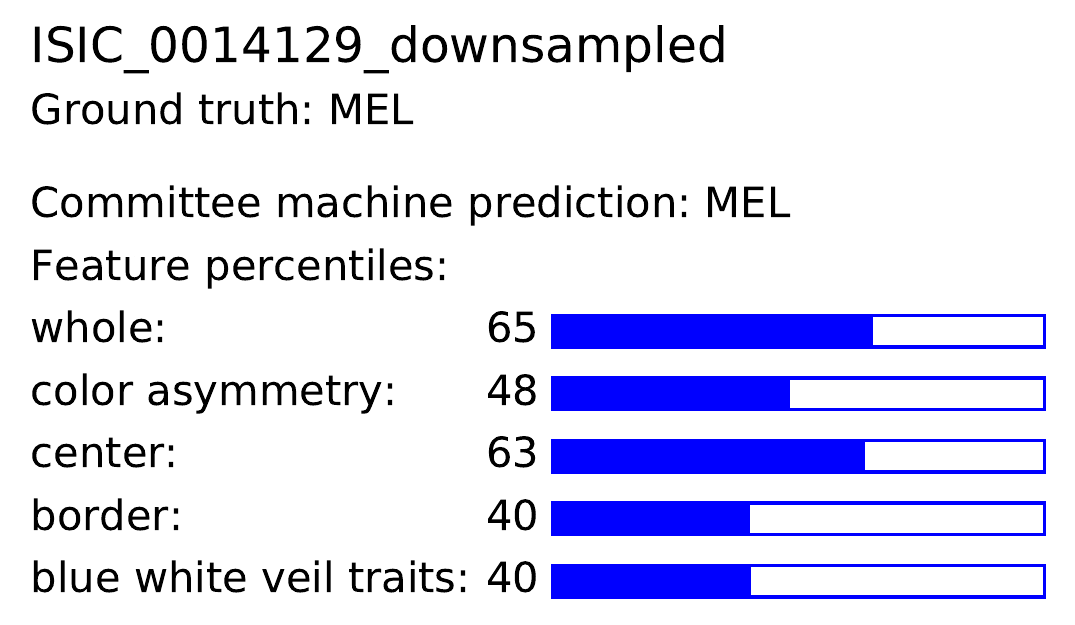}} & MEL (uncertain)\newline big size, red inflamed, regression can be seen & non-MEL (certain)\newline single color, symmetric pigmentation pattern \\
 \end{tabular}
\end{table}

\begin{table}[]
    \begin{tabular}{c c p{45mm} p{45mm}}
        Image & Ground truth and predictions & Expert 1 & Expert 2 \\
        \hline
        
\rule{0mm}{10mm}\raisebox{-28mm}{\includegraphics[height=30mm]{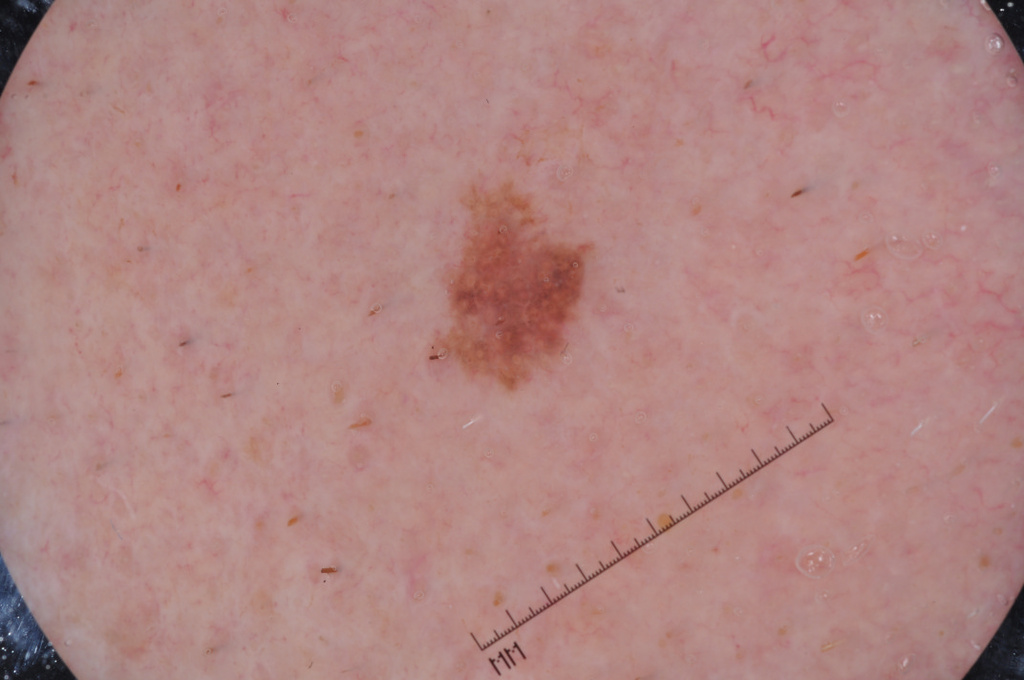}} & \raisebox{-26mm}{\includegraphics[height=28mm]{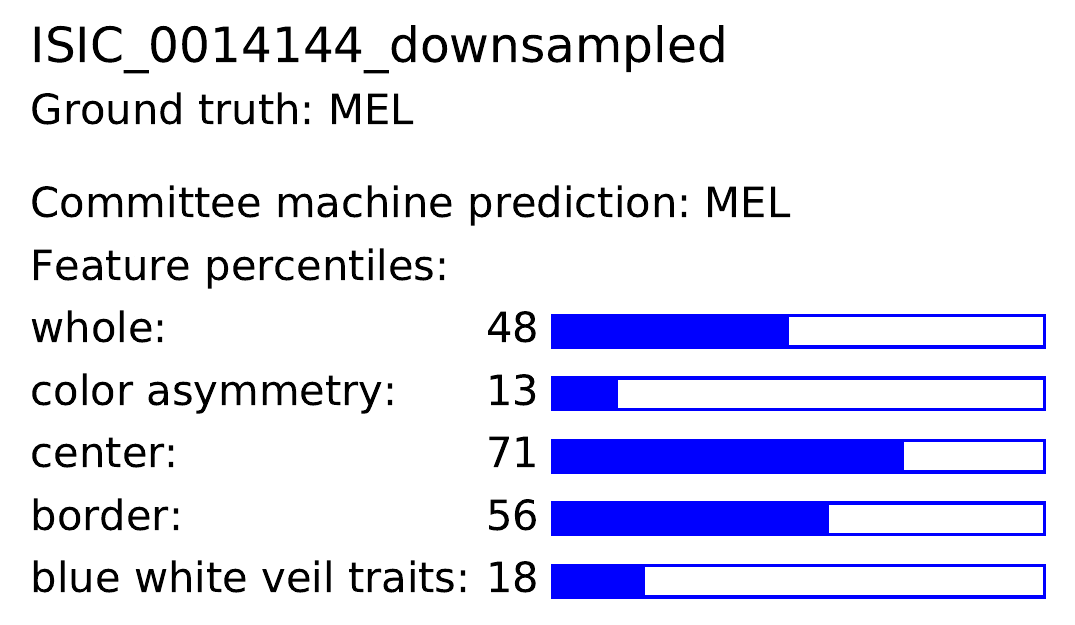}} & non-MEL (uncertain)\newline pale peppering, hard to judge & non-MEL (uncertain)\newline single color, symmetric pigmentation pattern \\
\rule{0mm}{10mm}\raisebox{-28mm}{\includegraphics[height=30mm]{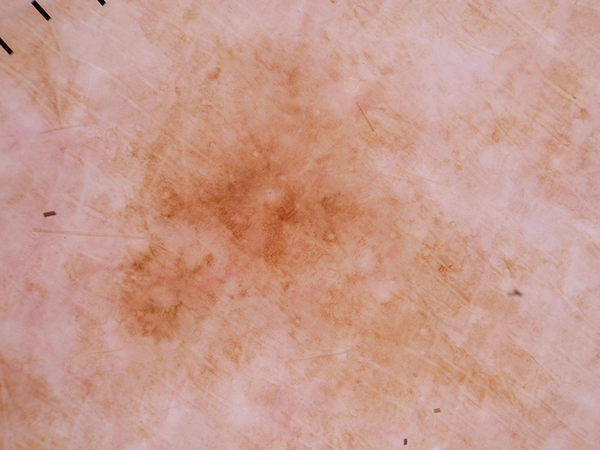}} & \raisebox{-26mm}{\includegraphics[height=28mm]{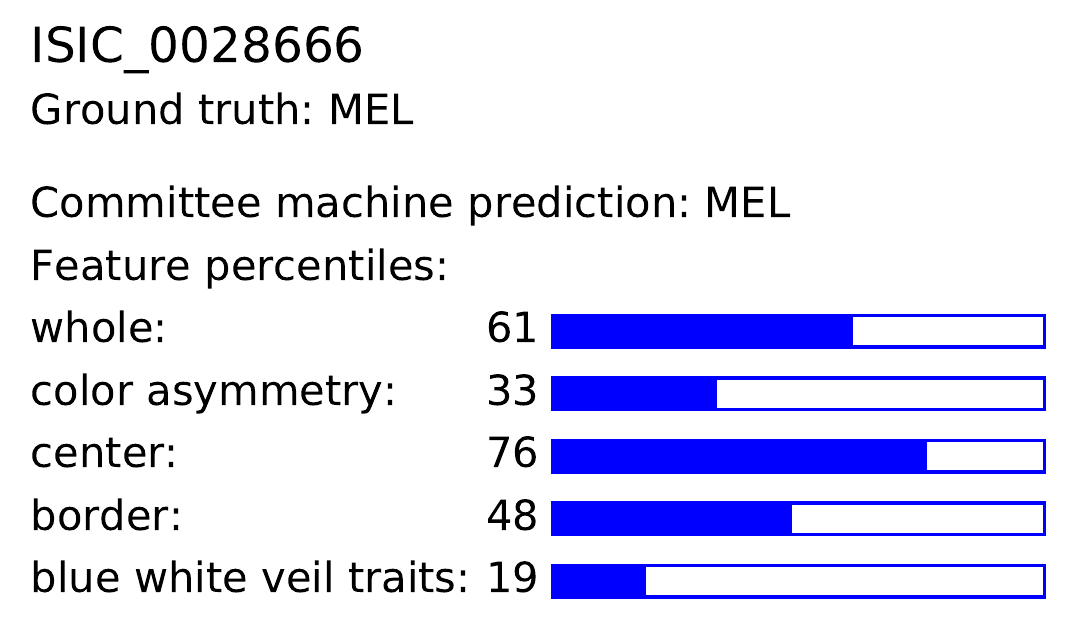}} & MEL (uncertain)\newline regression, single color,  & non-MEL (almost certain)\newline single color \\
\rule{0mm}{10mm}\raisebox{-28mm}{\includegraphics[height=30mm]{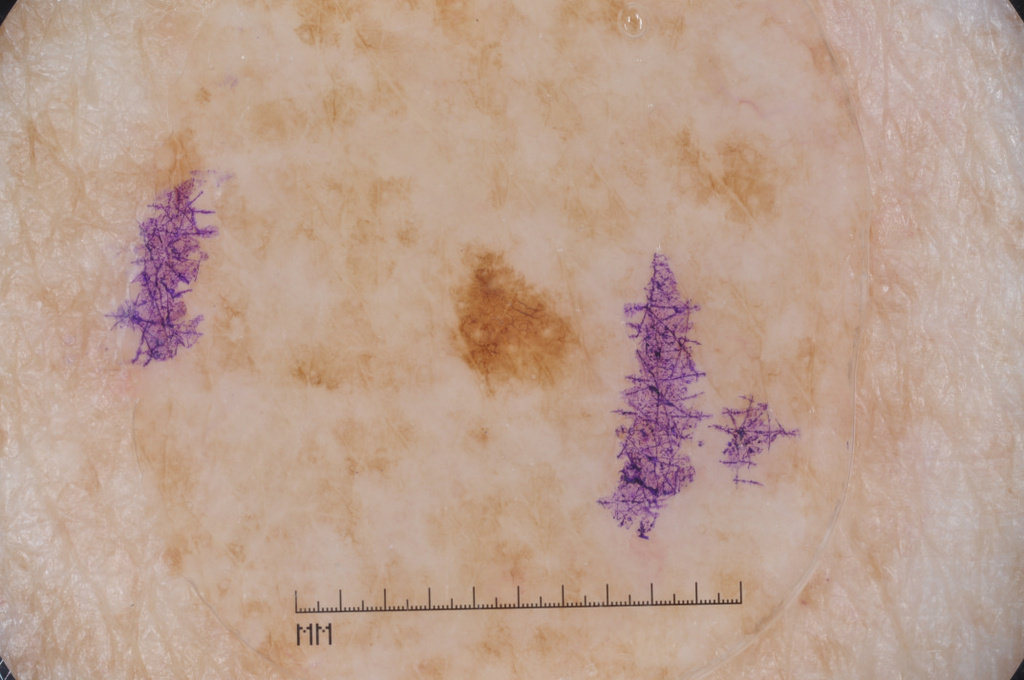}} & \raisebox{-26mm}{\includegraphics[height=28mm]{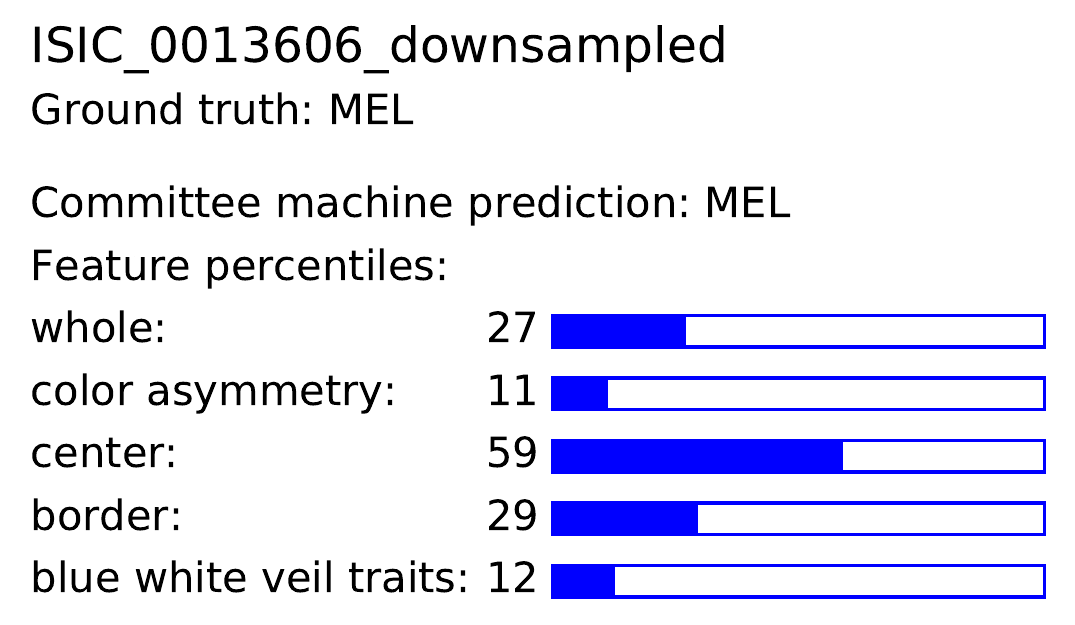}} & non-MEL (certain)\newline homogeneous but has darker and brighter parts & non-MEL (certain)\newline  \\
\rule{0mm}{10mm}\raisebox{-28mm}{\includegraphics[height=30mm]{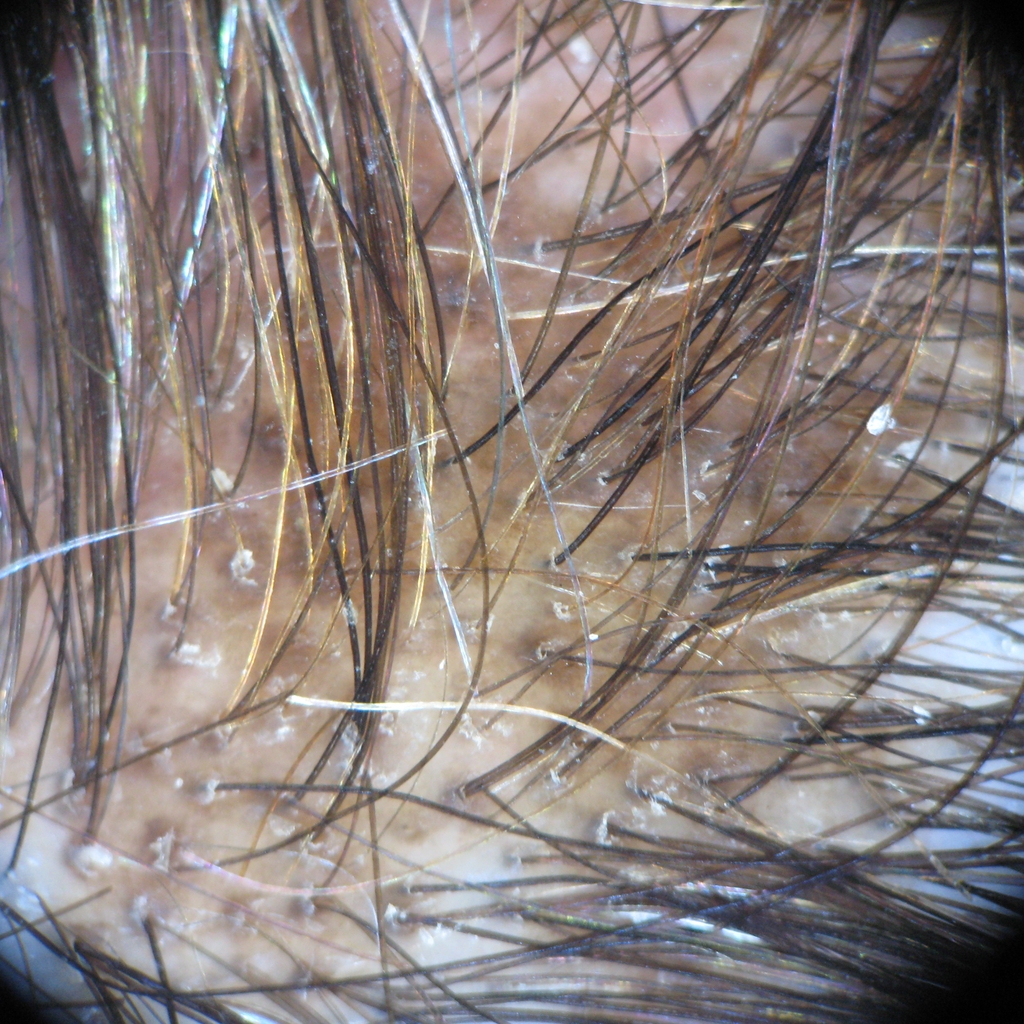}} & \raisebox{-26mm}{\includegraphics[height=28mm]{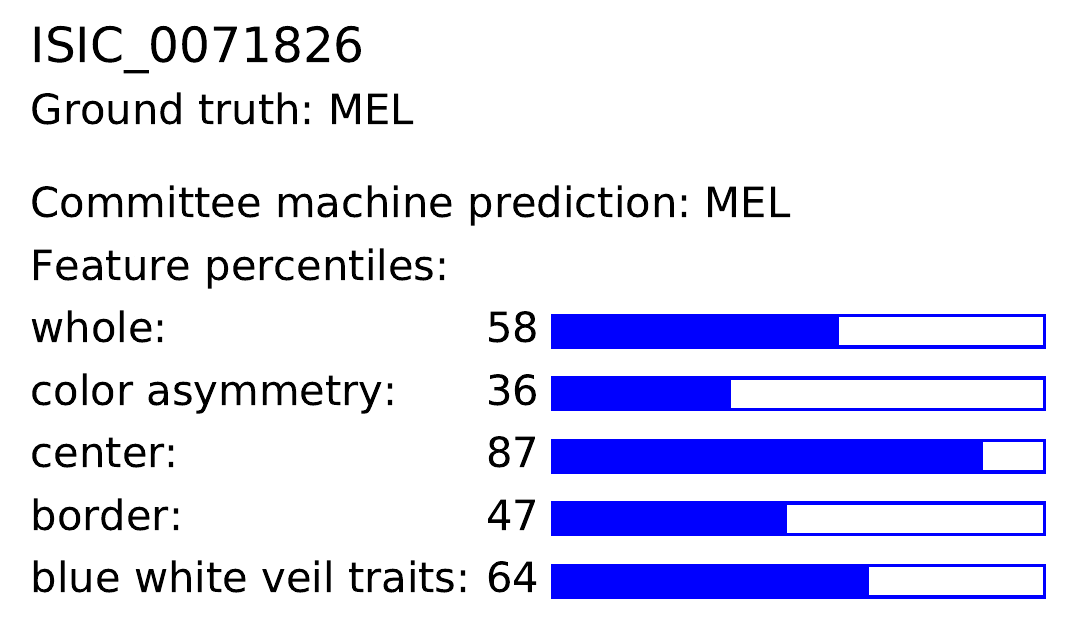}} & MEL (almost certain)\newline there could be a blue-white veil & non-MEL (uncertain)\newline  \\
\rule{0mm}{10mm}\raisebox{-28mm}{\includegraphics[height=30mm]{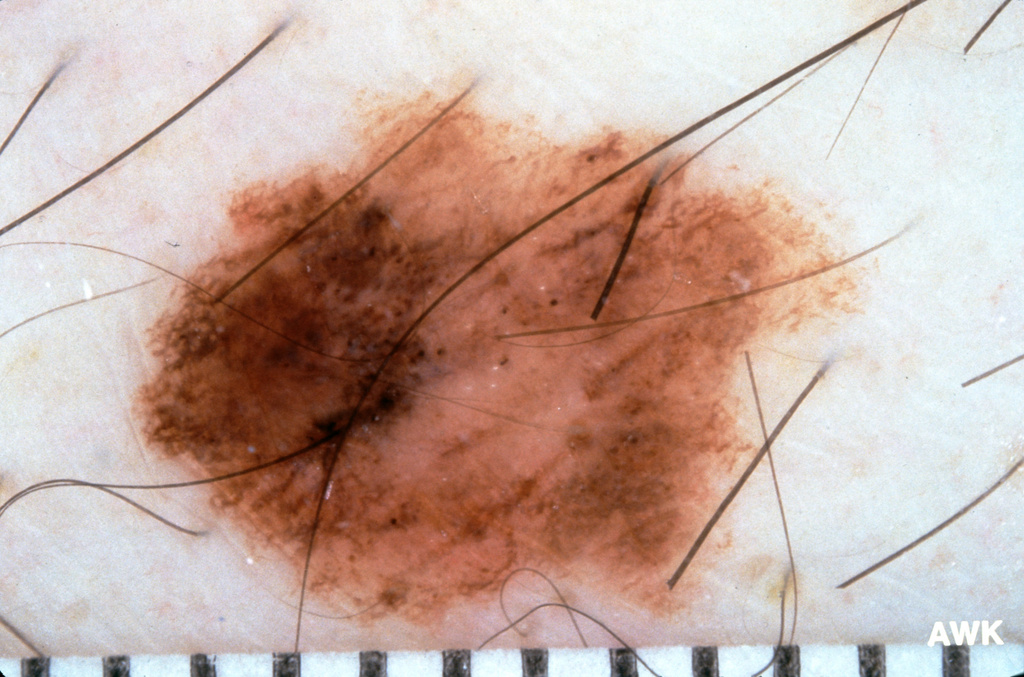}} & \raisebox{-26mm}{\includegraphics[height=28mm]{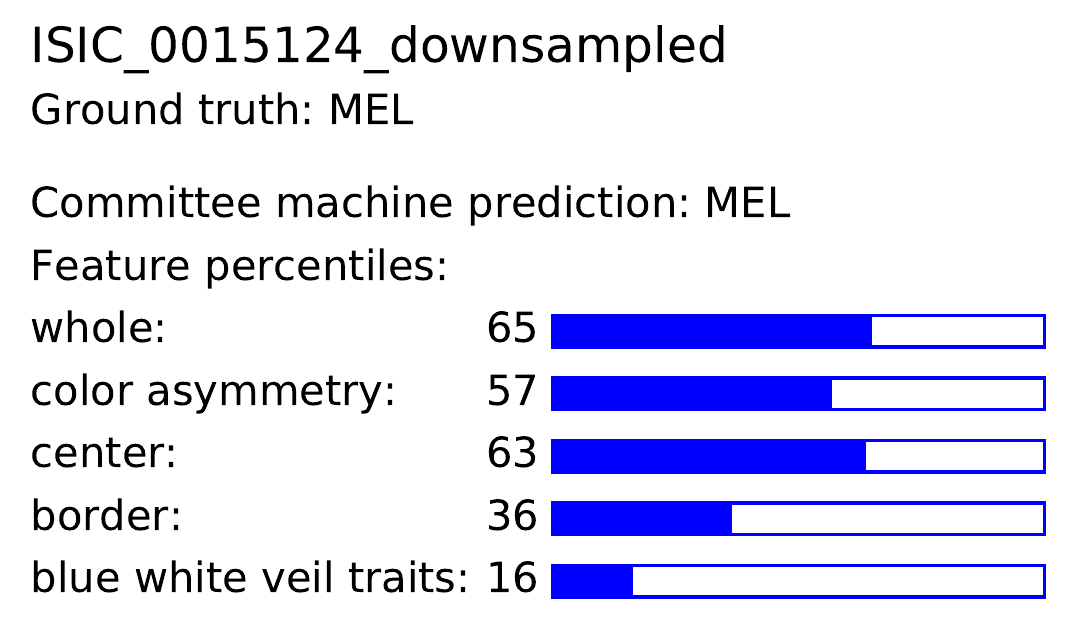}} & MEL (almost certain)\newline very asymmetrical lesion, black dots & MEL (uncertain)\newline broadened network, peripheral black dots/globules \\
 \end{tabular}
\end{table}

\begin{table}[]
    \begin{tabular}{c c p{45mm} p{45mm}}
        Image & Ground truth and predictions & Expert 1 & Expert 2 \\
        \hline
        
\rule{0mm}{10mm}\raisebox{-28mm}{\includegraphics[height=30mm]{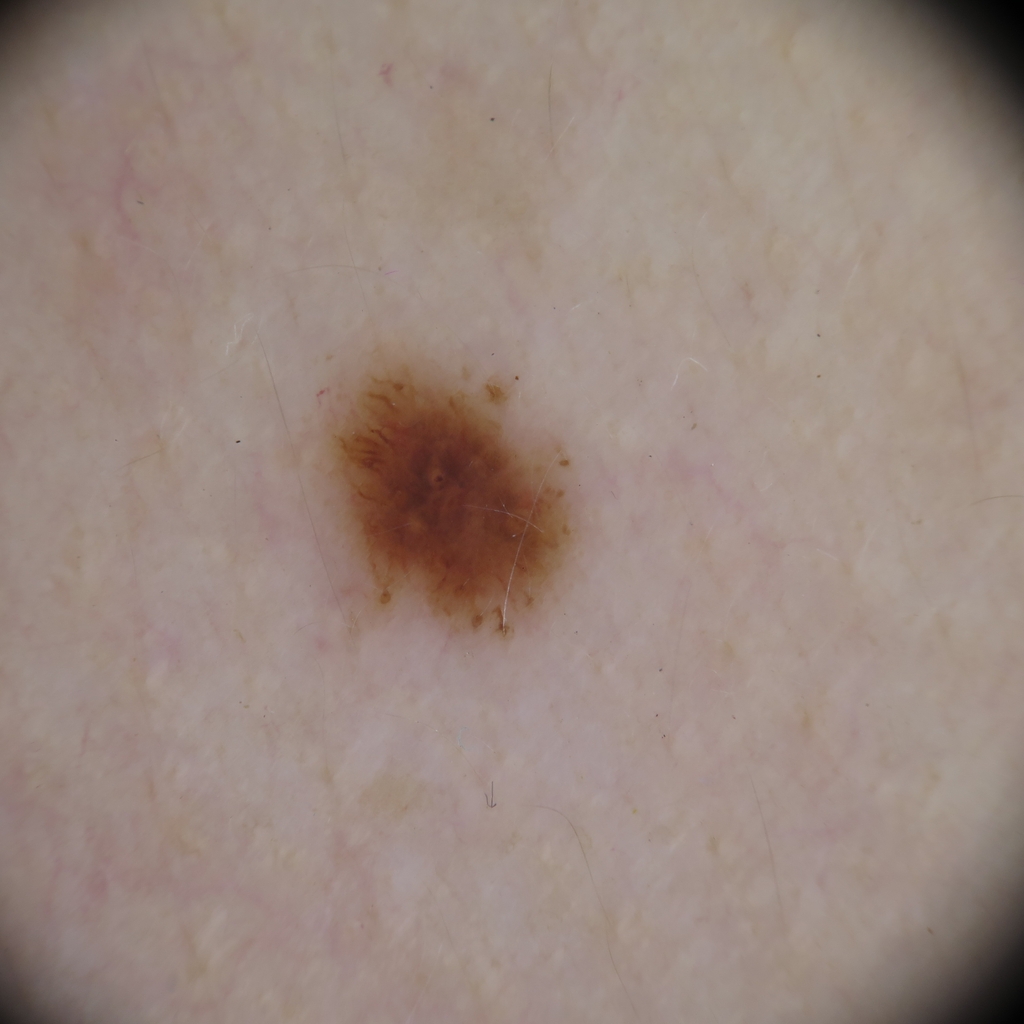}} & \raisebox{-26mm}{\includegraphics[height=28mm]{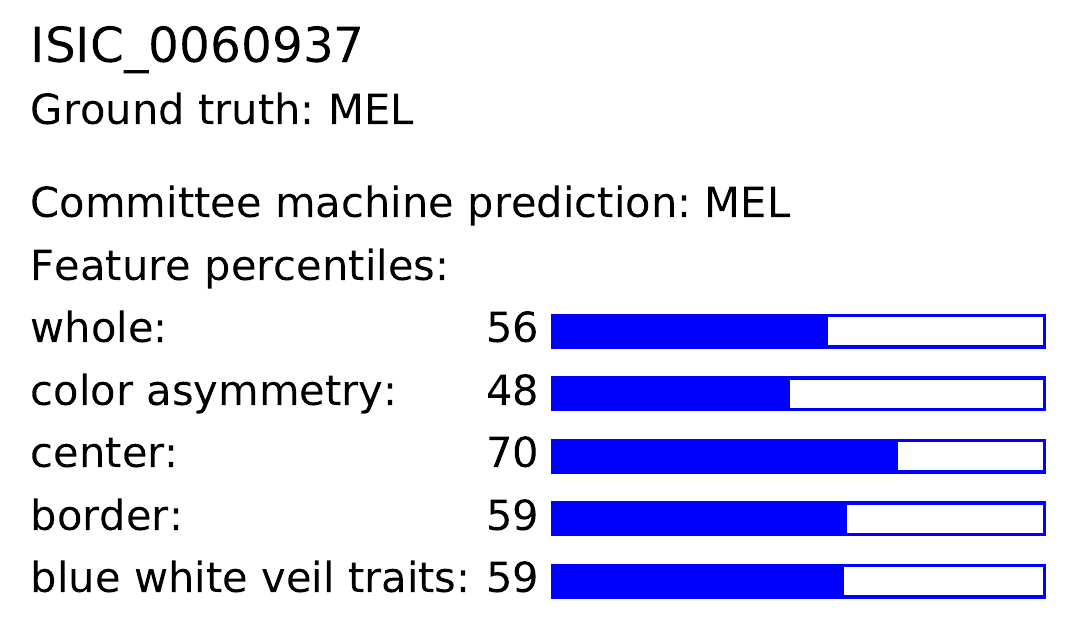}} & non-MEL (almost certain)\newline inflamed lesion, homogeneous, small radial streaming & non-MEL (uncertain)\newline single color, symmetric pigmentation pattern \\
\rule{0mm}{10mm}\raisebox{-28mm}{\includegraphics[height=30mm]{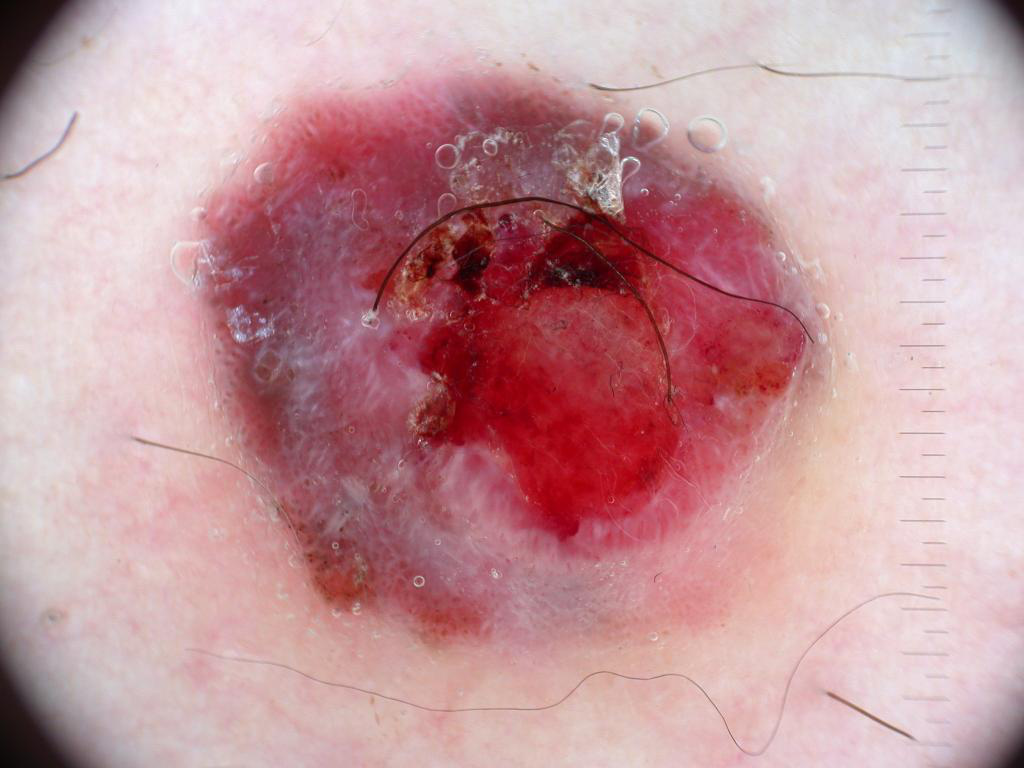}} & \raisebox{-26mm}{\includegraphics[height=28mm]{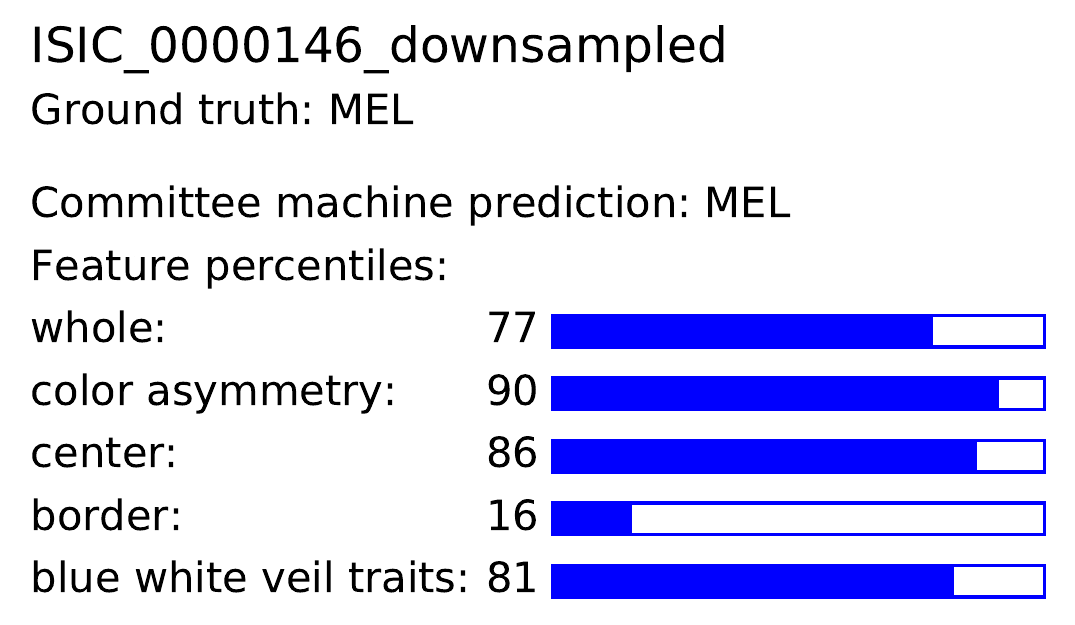}} & MEL (certain)\newline inflamed, ulcerous, nodular & MEL (certain)\newline 5-6 colors, scar-like depigmentation, pseudopods, multiple brown dots \\
\rule{0mm}{10mm}\raisebox{-28mm}{\includegraphics[height=30mm]{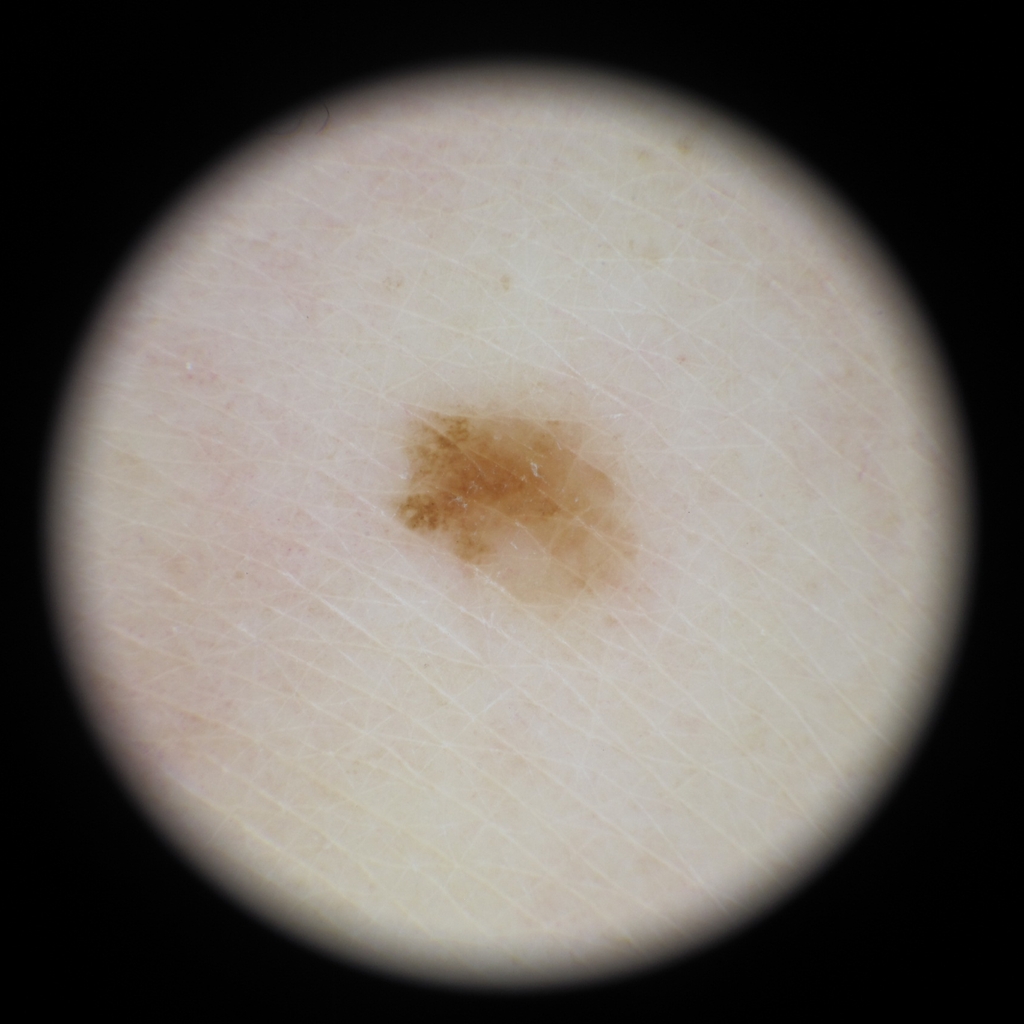}} & \raisebox{-26mm}{\includegraphics[height=28mm]{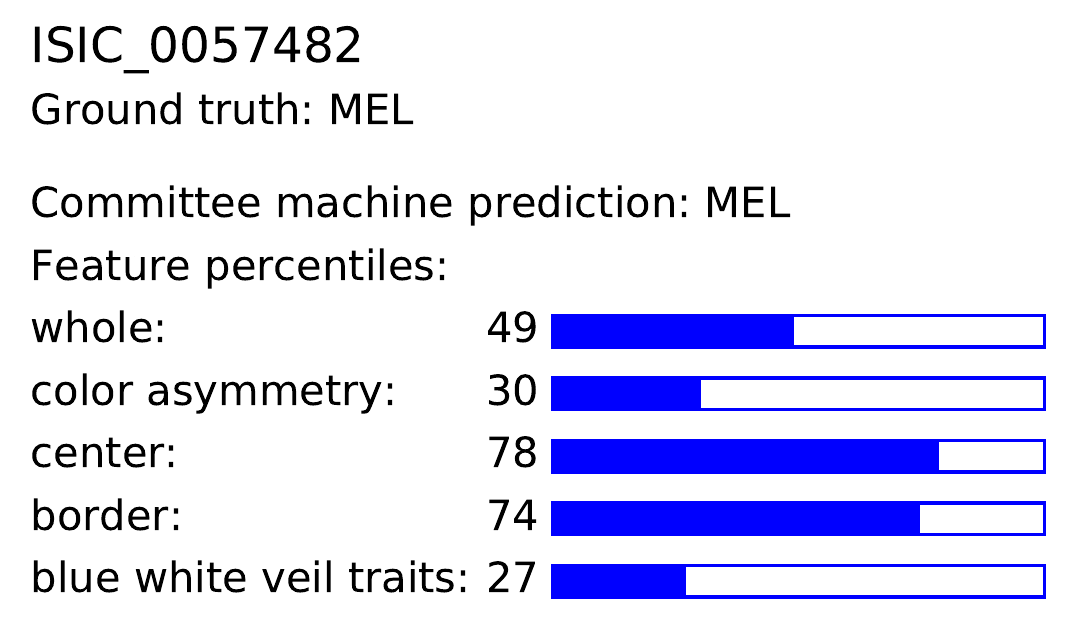}} & non-MEL (certain)\newline pale structures, homogeneous & non-MEL (certain)\newline  \\
\rule{0mm}{10mm}\raisebox{-28mm}{\includegraphics[height=30mm]{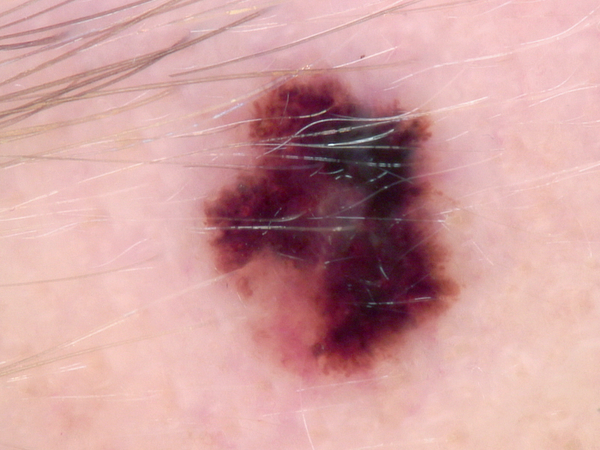}} & \raisebox{-26mm}{\includegraphics[height=28mm]{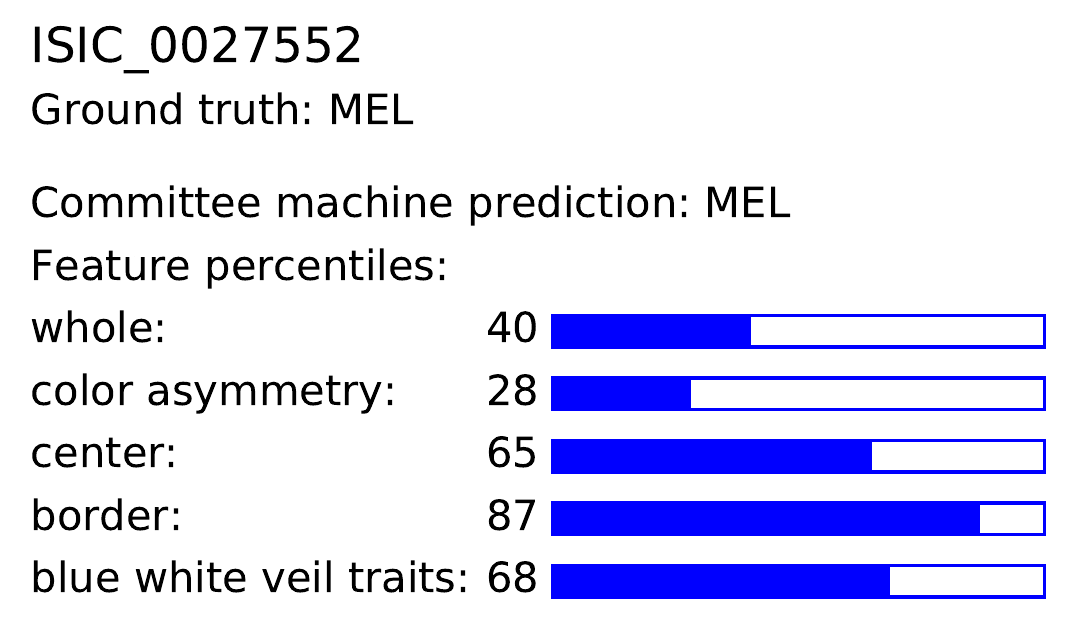}} & MEL (certain)\newline inflamed, structureless areas, blue-white veil, asymmetrical lesion, radial streaming & MEL (certain)\newline 5-6 colors, scar-like depigmentation, radial streaming, pseudopods, blue-white veil \\
\rule{0mm}{10mm}\raisebox{-28mm}{\includegraphics[height=30mm]{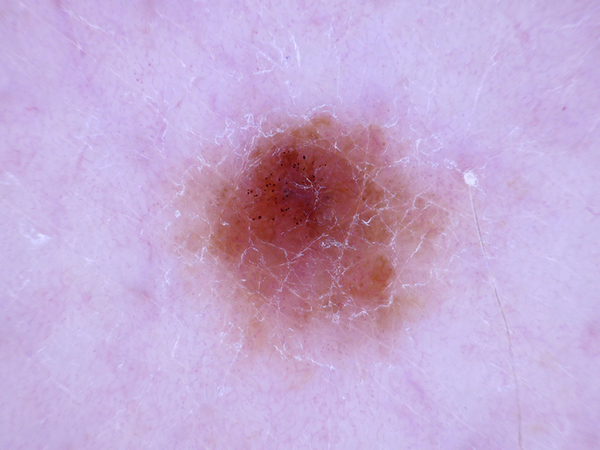}} & \raisebox{-26mm}{\includegraphics[height=28mm]{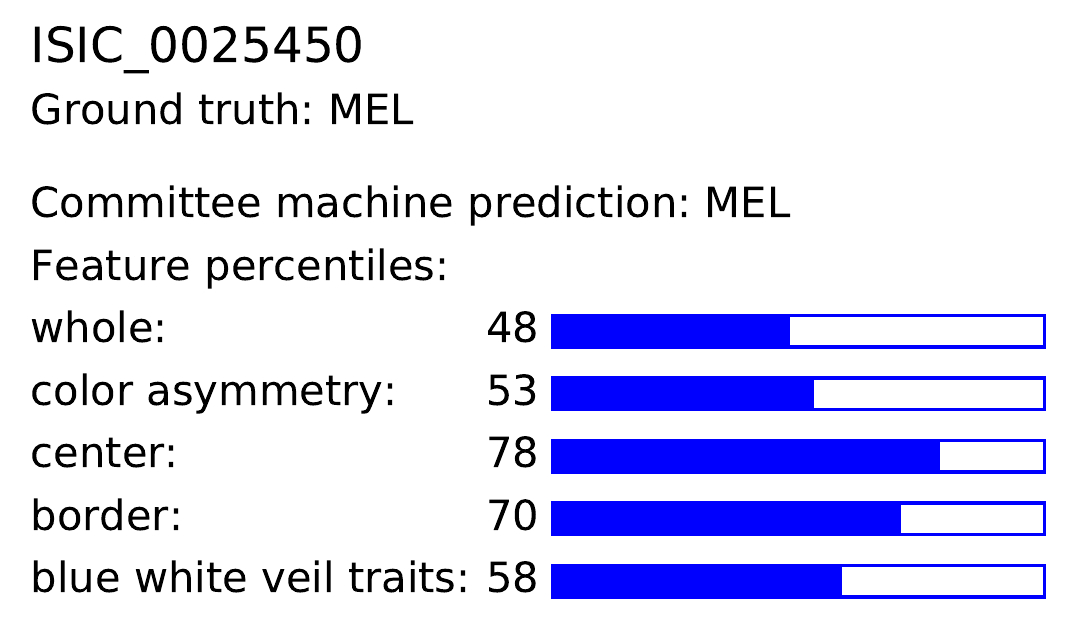}} & MEL (certain)\newline inflamed lesion, black dots and globules, blue-white veil & MEL (uncertain)\newline scar-like depigmentation, multiple brown dots, symmetric pigmentation pattern \\
 \end{tabular}
\end{table}

\begin{table}[]
    \begin{tabular}{c c p{45mm} p{45mm}}
        Image & Ground truth and predictions & Expert 1 & Expert 2 \\
        \hline
        
\rule{0mm}{10mm}\raisebox{-28mm}{\includegraphics[height=30mm]{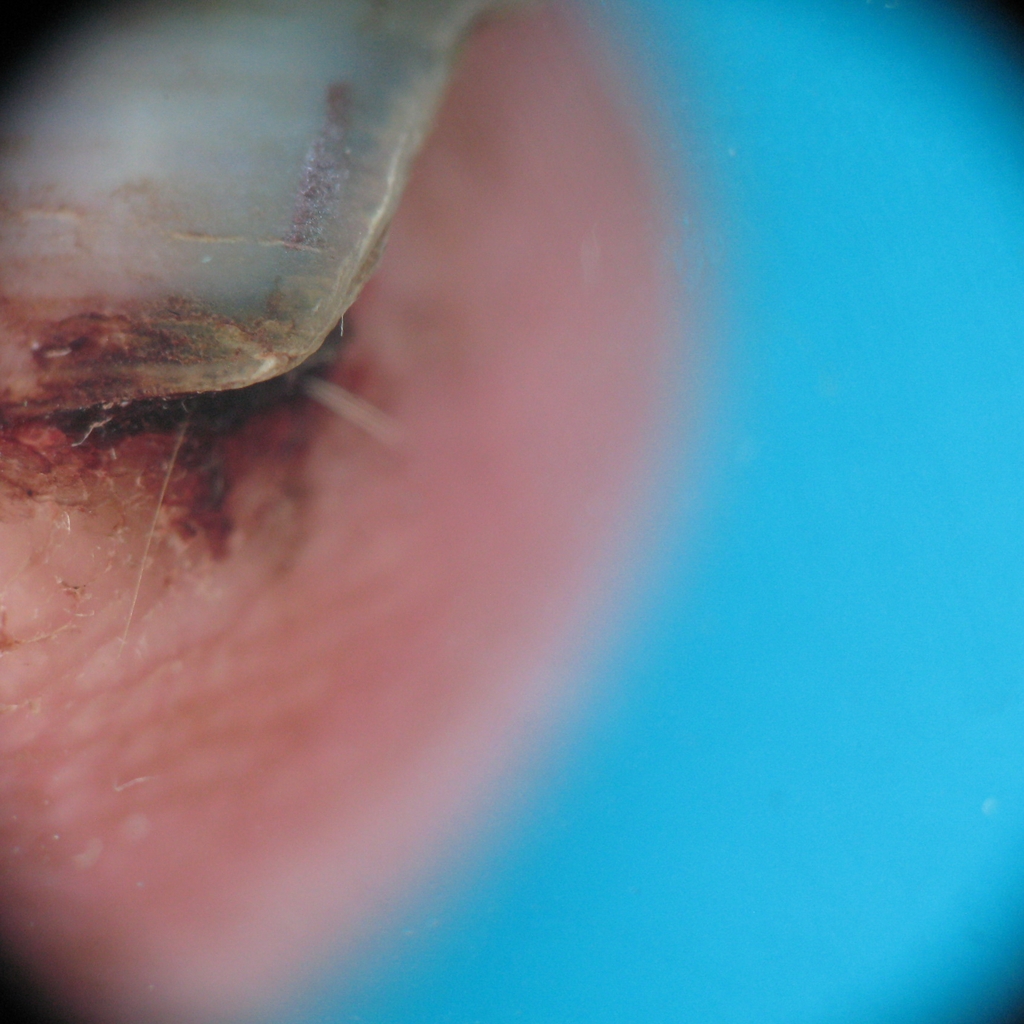}} & \raisebox{-26mm}{\includegraphics[height=28mm]{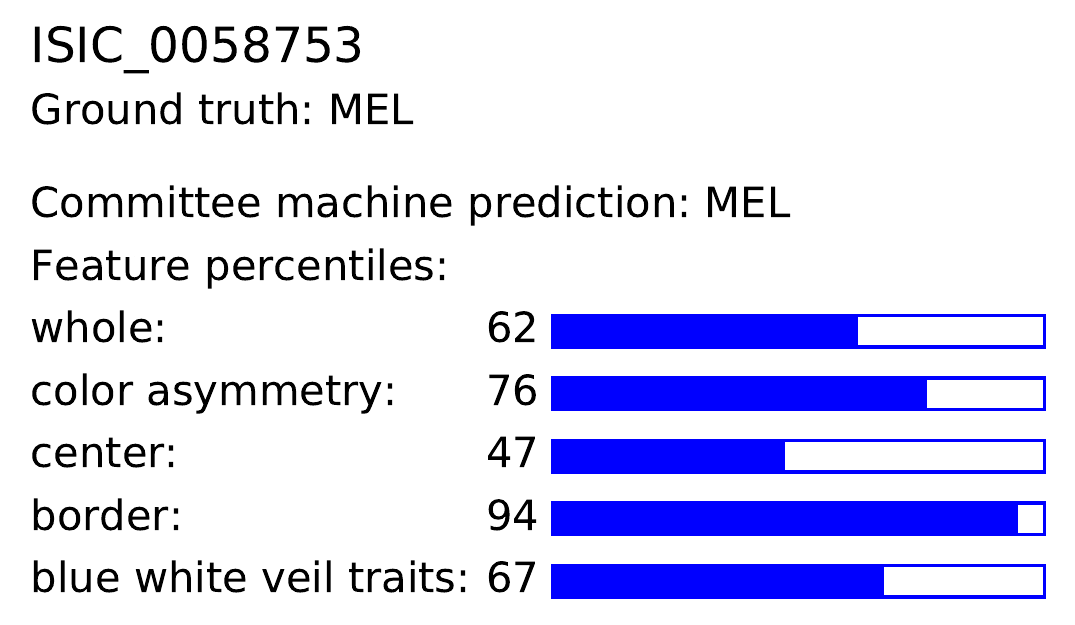}} &  (uncertain)\newline it could be only a haematoma but it is also possible that the melanoma is bleeding. It could not be judged from this image & MEL (uncertain)\newline  \\
\rule{0mm}{10mm}\raisebox{-28mm}{\includegraphics[height=30mm]{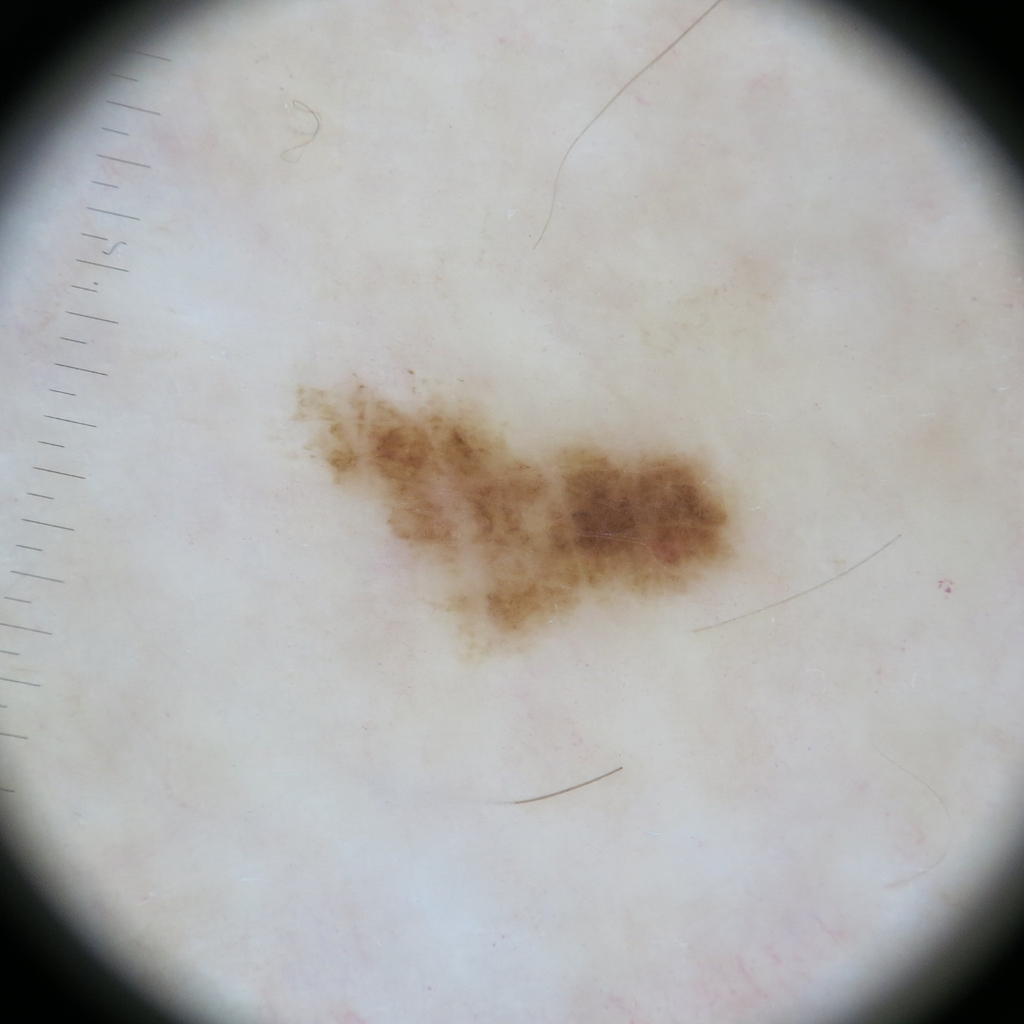}} & \raisebox{-26mm}{\includegraphics[height=28mm]{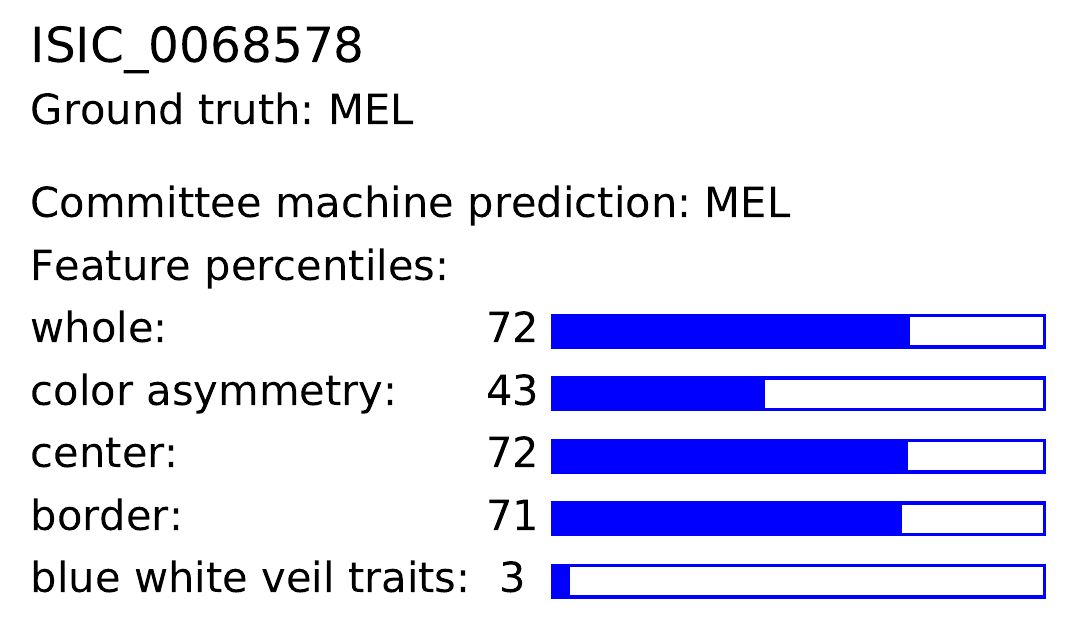}} & non-MEL (uncertain)\newline pale globules,  & non-MEL (almost certain)\newline single color, symmetric pigmentation pattern \\
\rule{0mm}{10mm}\raisebox{-28mm}{\includegraphics[height=30mm]{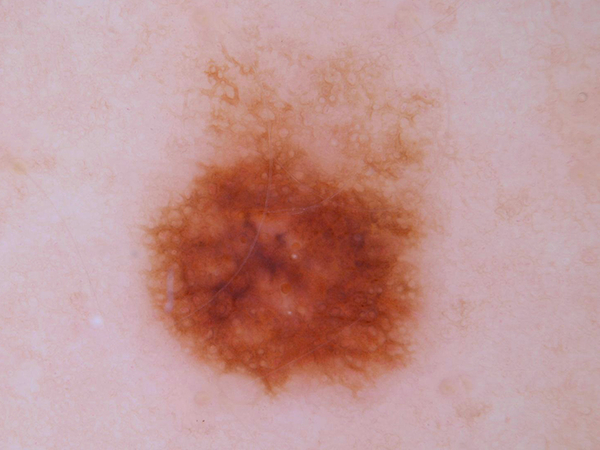}} & \raisebox{-26mm}{\includegraphics[height=28mm]{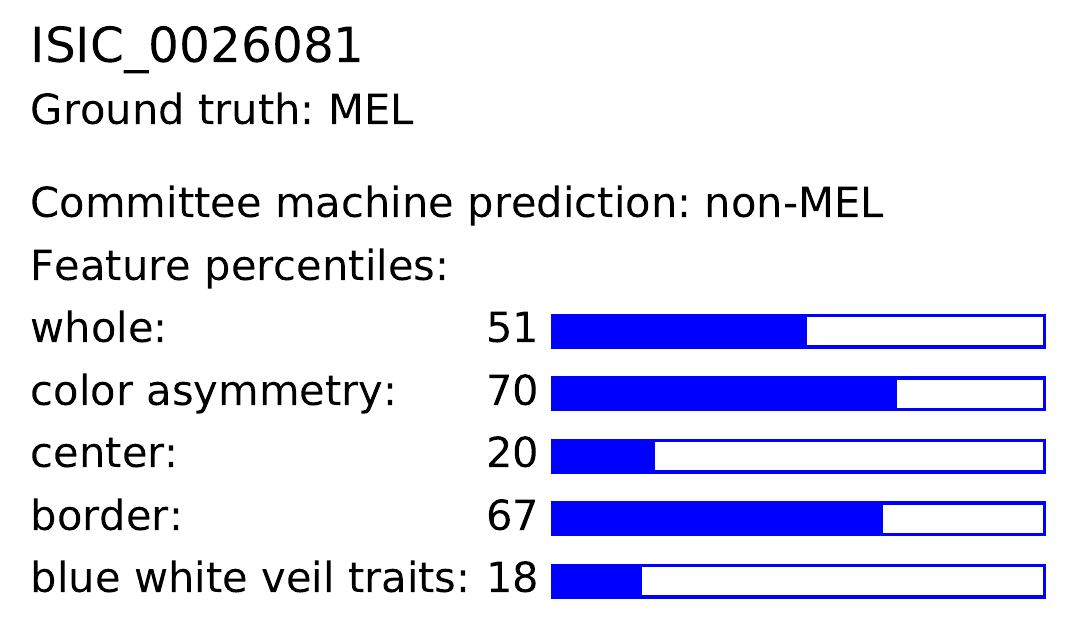}} & MEL (uncertain)\newline blue-white veil, different upper and lower structure, dots and globules & non-MEL (almost certain)\newline single color,  \\
\rule{0mm}{10mm}\raisebox{-28mm}{\includegraphics[height=30mm]{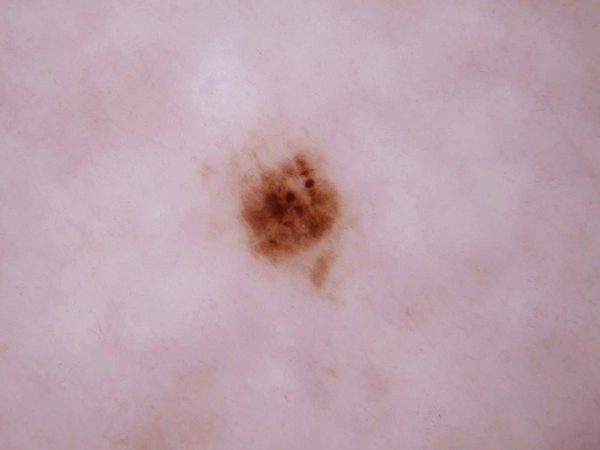}} & \raisebox{-26mm}{\includegraphics[height=28mm]{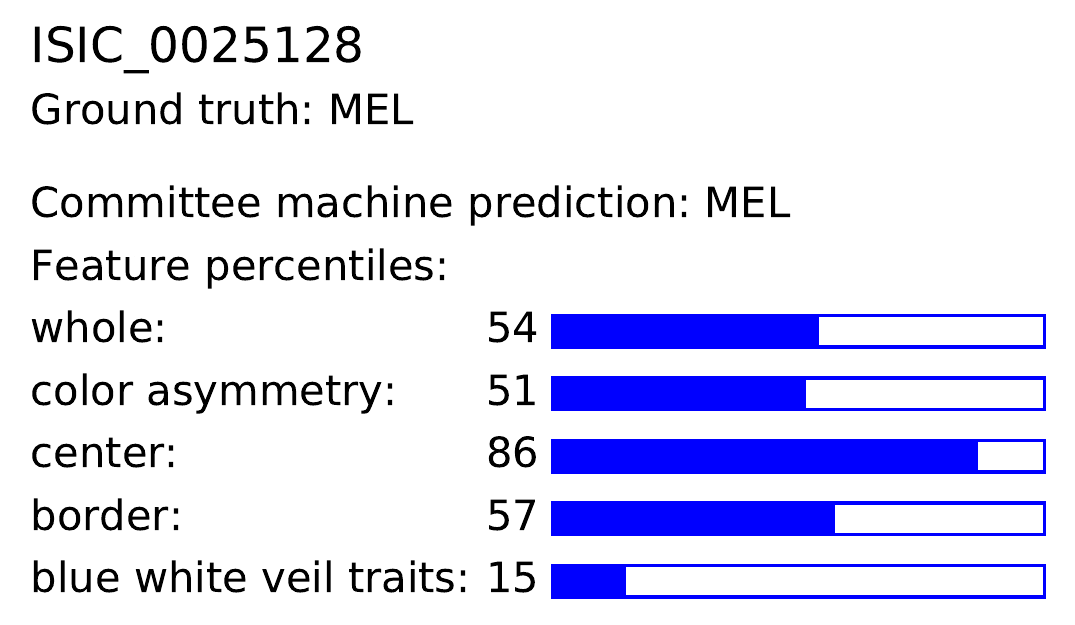}} & MEL (uncertain)\newline a satellite lesion can be seen, radial streaming & non-MEL (almost certain)\newline single color, symmetric pigmentation pattern \\
\rule{0mm}{10mm}\raisebox{-28mm}{\includegraphics[height=30mm]{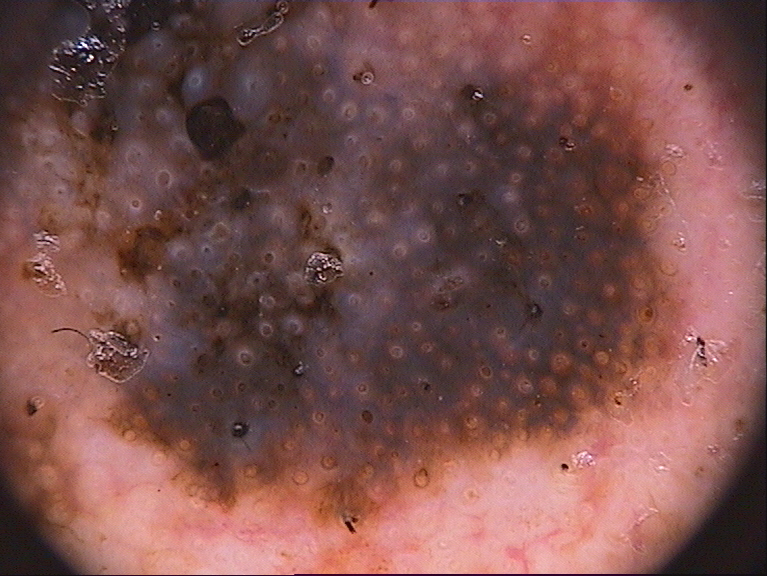}} & \raisebox{-26mm}{\includegraphics[height=28mm]{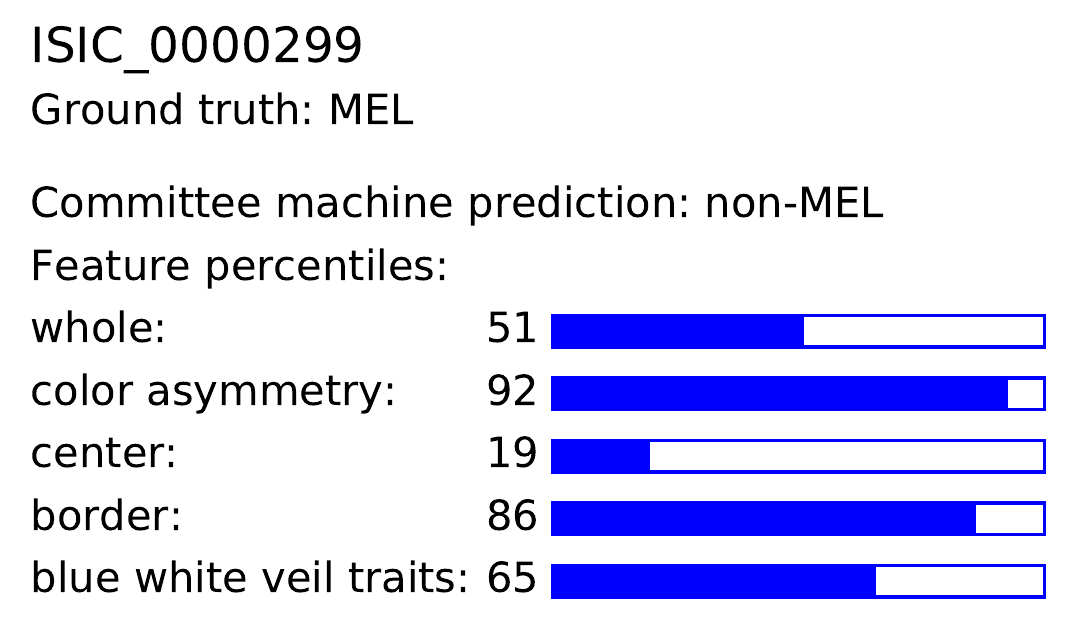}} & MEL (certain)\newline asymmetrical lesion, dark blue-black colors, veil & MEL (certain)\newline radial streaming, blue-white veil \\
 \end{tabular}
\end{table}

\begin{table}[]
    \begin{tabular}{c c p{45mm} p{45mm}}
        Image & Ground truth and predictions & Expert 1 & Expert 2 \\
        \hline
        
\rule{0mm}{10mm}\raisebox{-28mm}{\includegraphics[height=30mm]{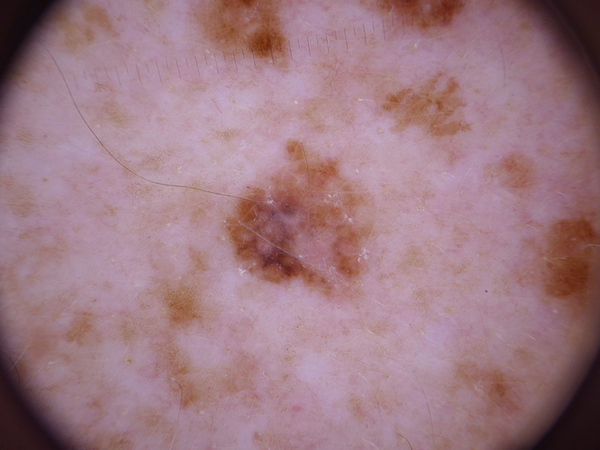}} & \raisebox{-26mm}{\includegraphics[height=28mm]{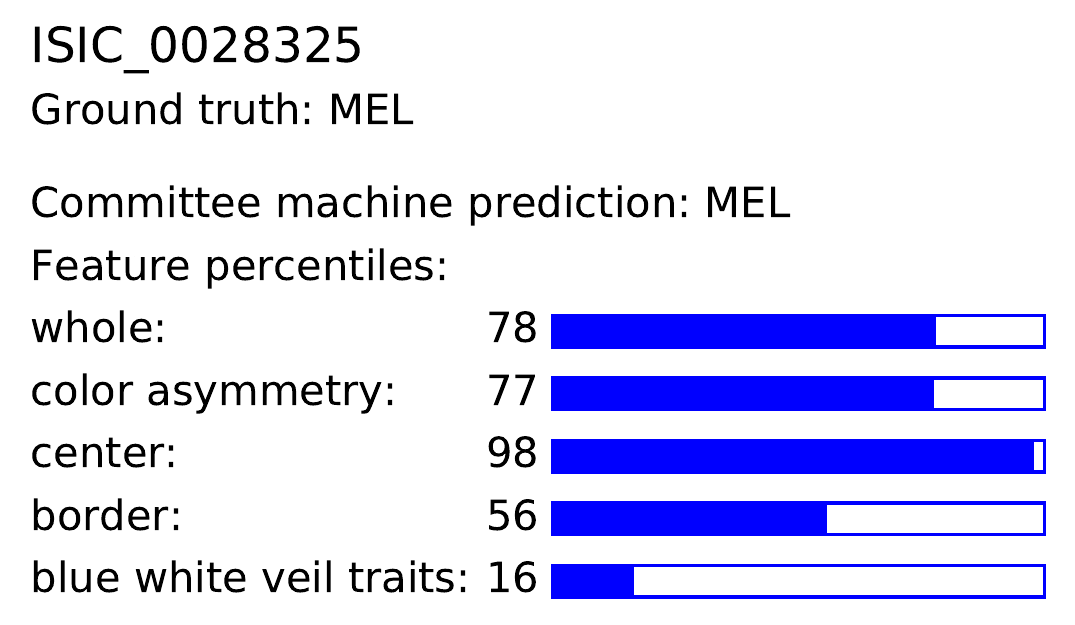}} & MEL (uncertain)\newline 3-4 colors, regression area & MEL (almost certain)\newline multiple blue-gray dots, 5-6 colors, scar-like depigmentation, blue-white veil \\
\rule{0mm}{10mm}\raisebox{-28mm}{\includegraphics[height=30mm]{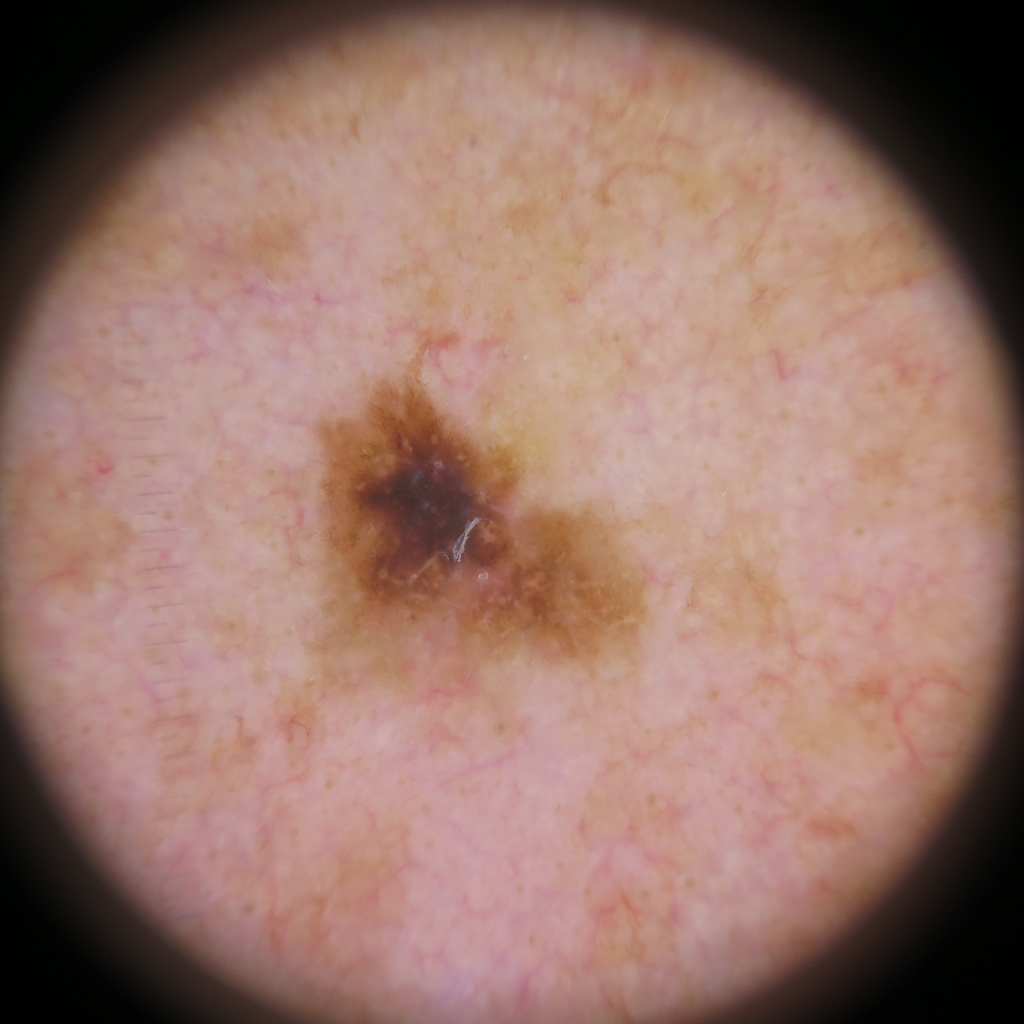}} & \raisebox{-26mm}{\includegraphics[height=28mm]{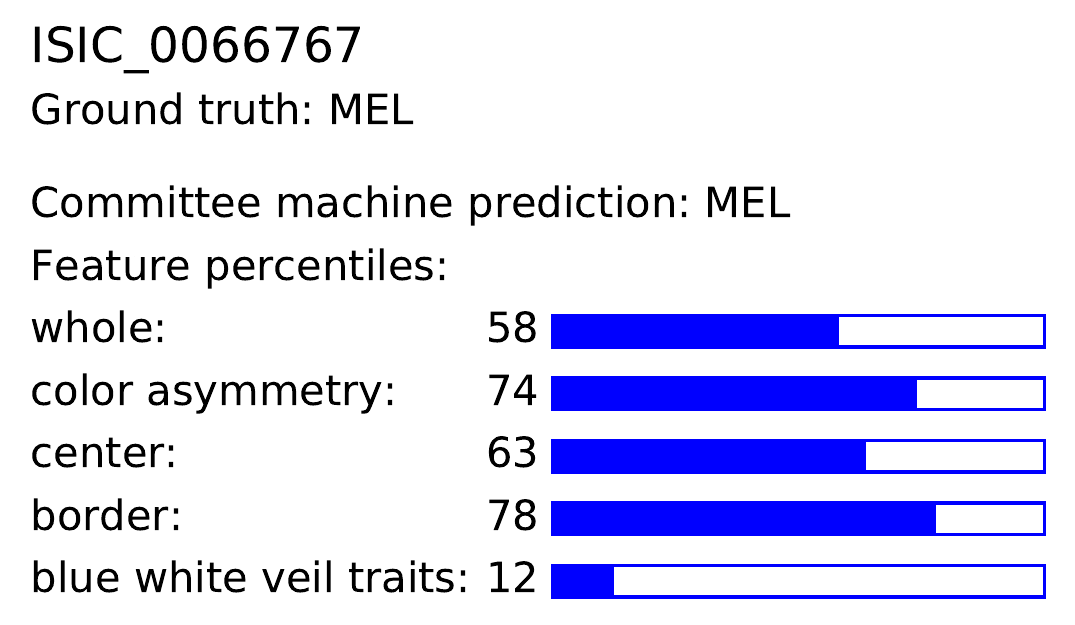}} & MEL (uncertain)\newline blue-white veil, radial streaming & MEL (uncertain)\newline 5-6 colors, scar-like depigmentation, radial streaming, blue-white veil \\
\rule{0mm}{10mm}\raisebox{-28mm}{\includegraphics[height=30mm]{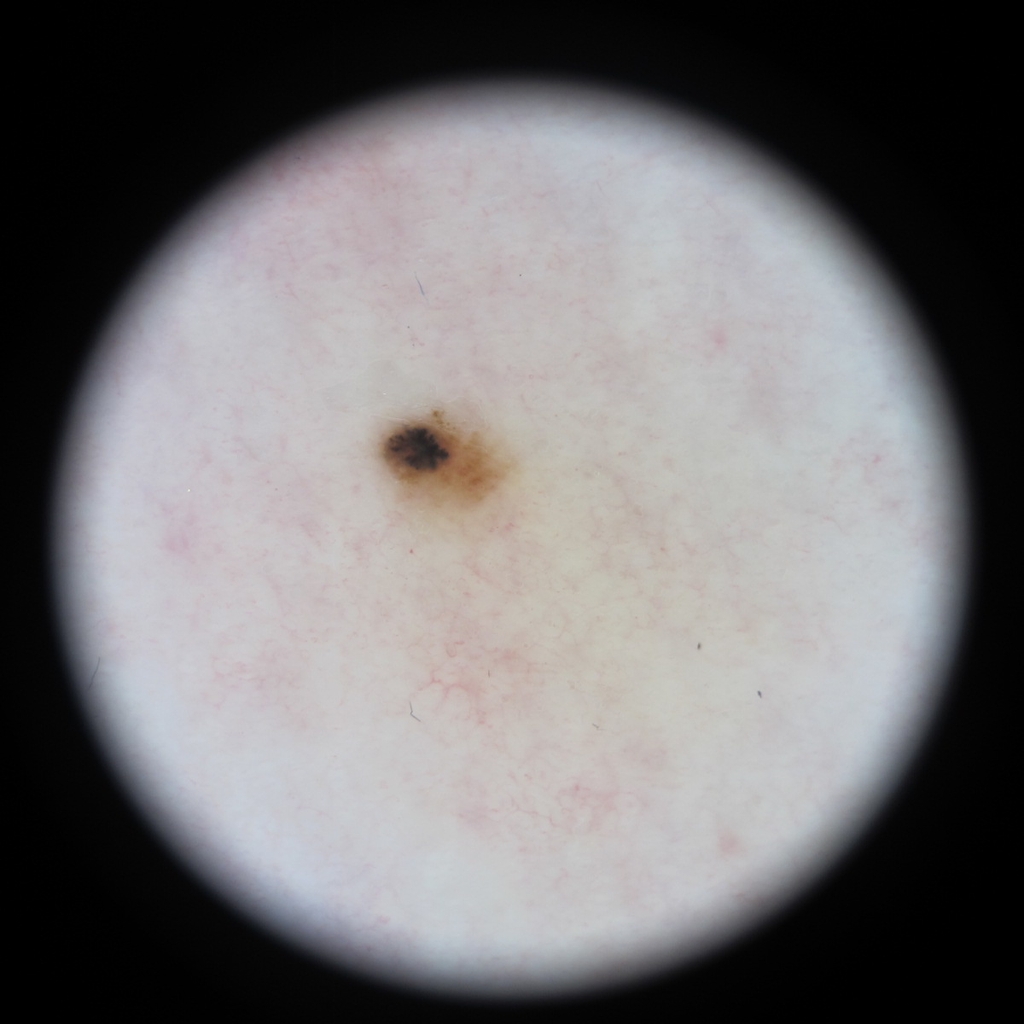}} & \raisebox{-26mm}{\includegraphics[height=28mm]{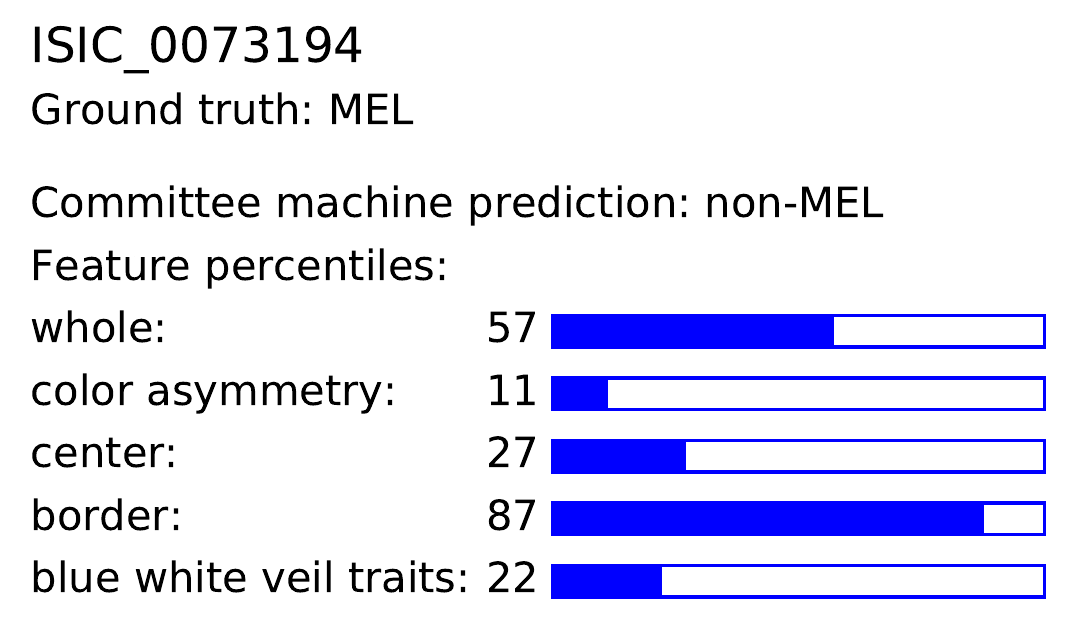}} & MEL (almost certain)\newline asymmetrical lesion, very small lesion, dark globules & non-MEL (uncertain)\newline  \\
\rule{0mm}{10mm}\raisebox{-28mm}{\includegraphics[height=30mm]{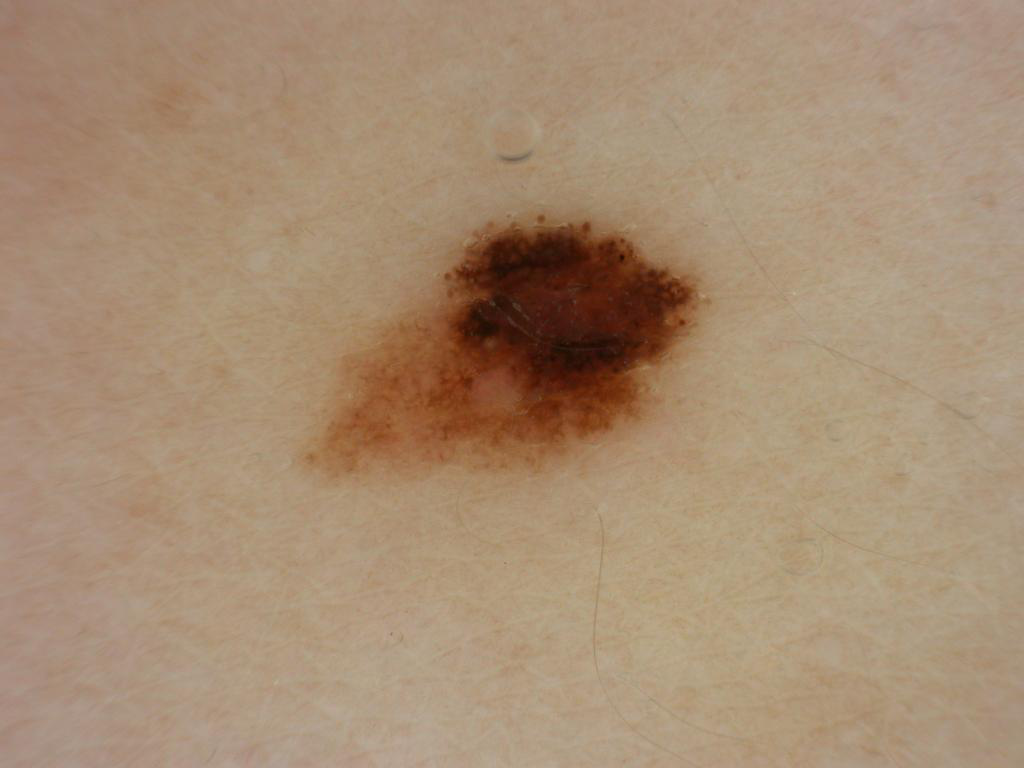}} & \raisebox{-26mm}{\includegraphics[height=28mm]{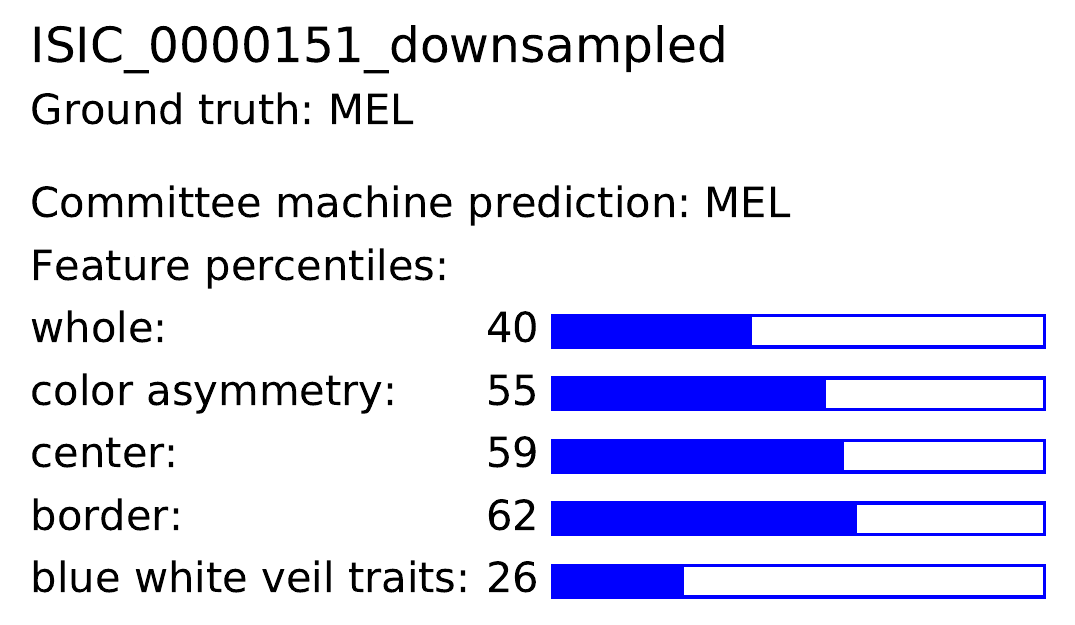}} & non-MEL (almost certain)\newline globules on the lesion's periphery, red basetone, a bit inflamed, it could be atypical nevus & MEL (almost certain)\newline peripheral black dots/globules, scar-like depigmentation, radial streaming, multiple brown dots \\
\rule{0mm}{10mm}\raisebox{-28mm}{\includegraphics[height=30mm]{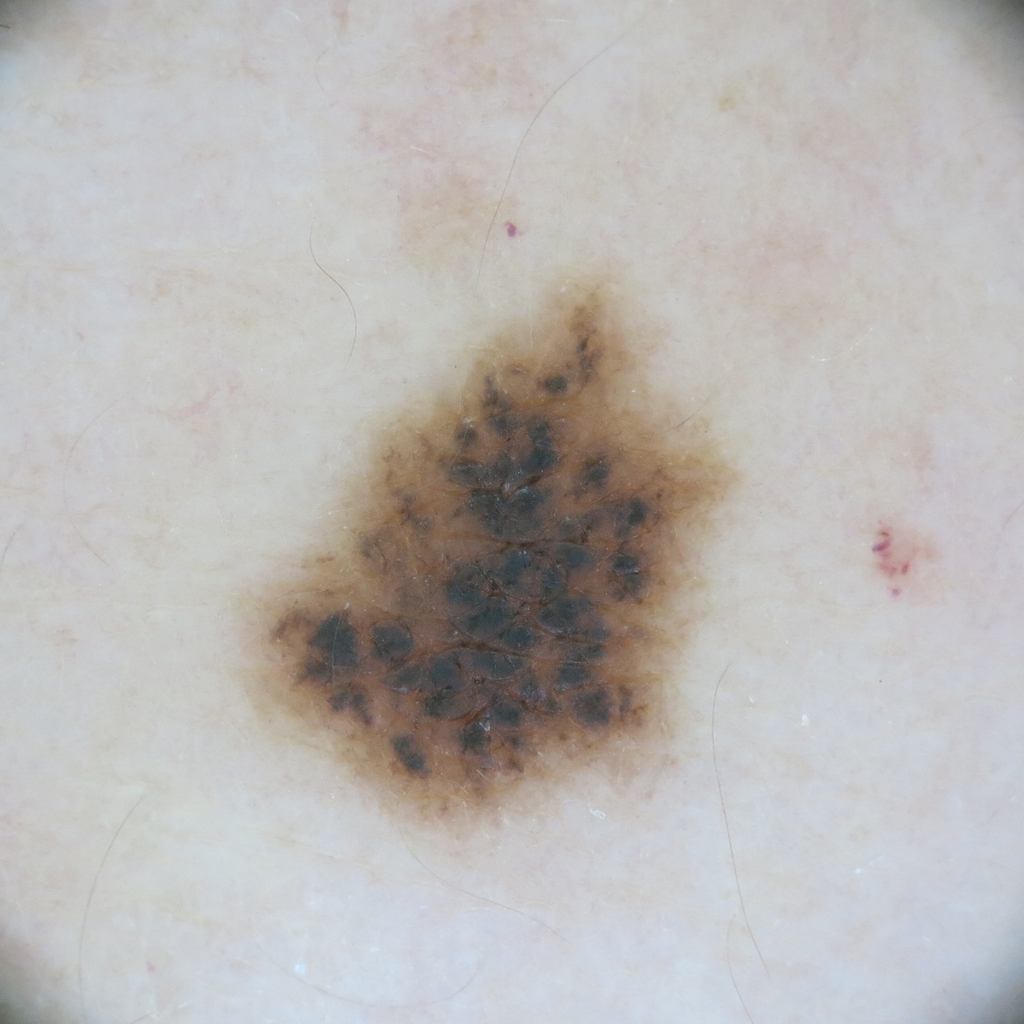}} & \raisebox{-26mm}{\includegraphics[height=28mm]{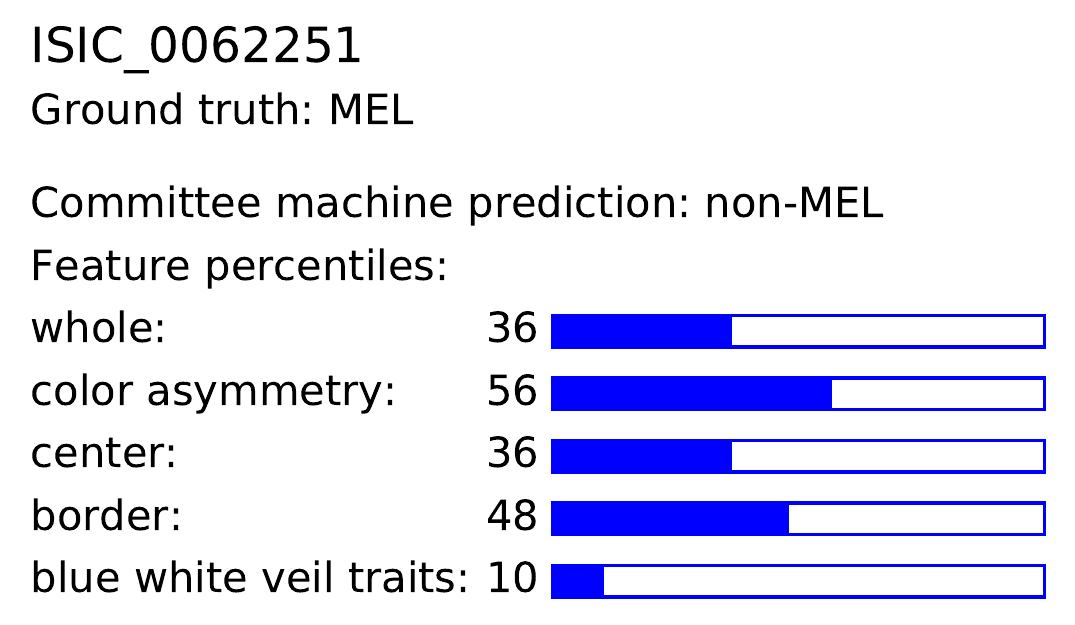}} & MEL (uncertain)\newline homogeneous lesion, suspicious black islands & MEL (uncertain)\newline  \\
 \end{tabular}
    \caption{\tablecaption}
\end{table}


\clearpage
\section{DenseNet169 results}

For comparison we repeated the analysis of the main paper with a weaker backbone. The results are in Table 2-5 and Fig.~1 below, the conclusions are in the main text.

\begin{table}[h]
    \centering
    \begin{tabular}{lcccc}
    
feature classifier & accuracy & balanced accuracy & false negative rate & ROC AUC \\
\hline
whole & $0.8975 \pm 0.0187$ & $0.8305 \pm 0.0160$ & $0.2738 \pm 0.0241$ & $0.9281 \pm 0.0120$\\
color asymmetry & $0.7984 \pm 0.0141$ & $0.7579 \pm 0.0041$ & $0.3052 \pm 0.0254$ & $0.8466 \pm 0.0034$\\
center & $0.8677 \pm 0.0112$ & $0.7366 \pm 0.0448$ & $0.4673 \pm 0.1046$ & $0.8816 \pm 0.0138$\\
border & $0.8533 \pm 0.0157$ & $0.7133 \pm 0.0247$ & $0.5044 \pm 0.0730$ & $0.8519 \pm 0.0181$\\
blue white veil traits$^*$ & $0.7635 \pm 0.0033$ & $0.6274 \pm 0.0083$ & $0.5843 \pm 0.0165$ & $0.6681 \pm 0.0129$\\

    \end{tabular}
    \medskip
    \caption{Benchmark results for the individual feature classifiers (DenseNet169 backbone). The displayed values and the uncertainties are the average and the standard deviation obtained by a 5-fold cross validation test.
    The accuracy, balanced accuracy and false negative rate are taken at the 0.5 threshold level.\newline
    $^*$The blue white veil traits detector, unlike the others, is trained for the target of blue white veil presence, and evaluated here for melanoma.}
    \label{tab:results-feature}
\end{table}

\begin{table}[h]
    \centering
    \begin{tabular}{lccc}
    
com. mach. input&accuracy&balanced accuracy&false negative rate\\
\hline
softmax&$0.9074\pm 0.0096$&$0.8375\pm 0.0176$&$0.2712\pm 0.0433$\\
feature layer&$0.9111\pm 0.0106$&$0.8422\pm 0.0165$&$0.2650\pm 0.0279$\\

    \end{tabular}
    \medskip
    \caption{Benchmark results obtained by the committee machine at the 0.5 threshold level for different input data (DenseNet169 backbone). The displayed values and uncertainties are the average and standard deviation obtained by a 5-fold cross validation test. To further improve statistics, for each fold 5 independent training of the committee machine took place starting from a different set of random weights.}
    \label{tab:committee}
\end{table}

\begin{figure}[ht]
    \centering
    \includegraphics[width=0.5\textwidth]{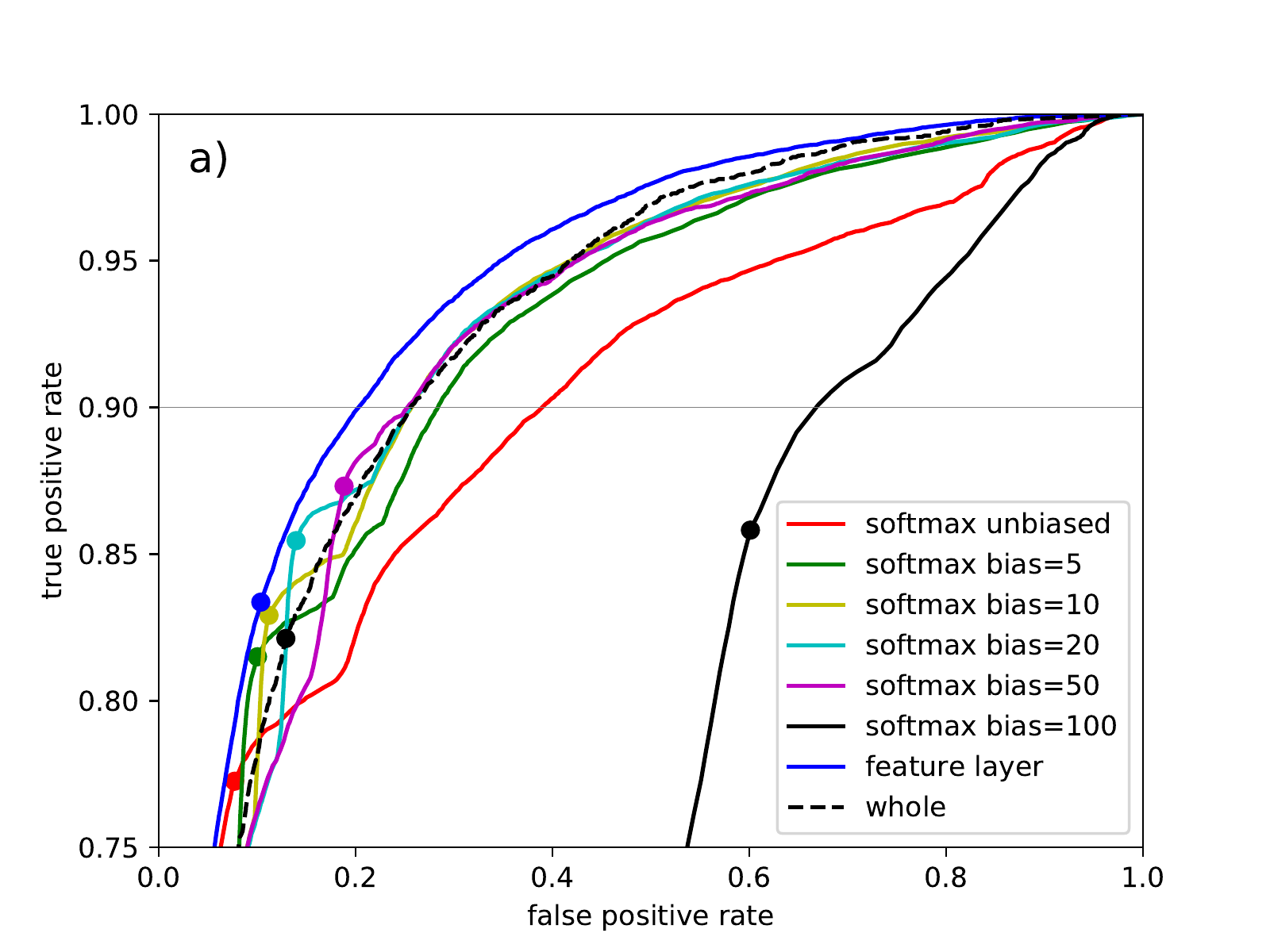}%
    \includegraphics[width=0.5\textwidth]{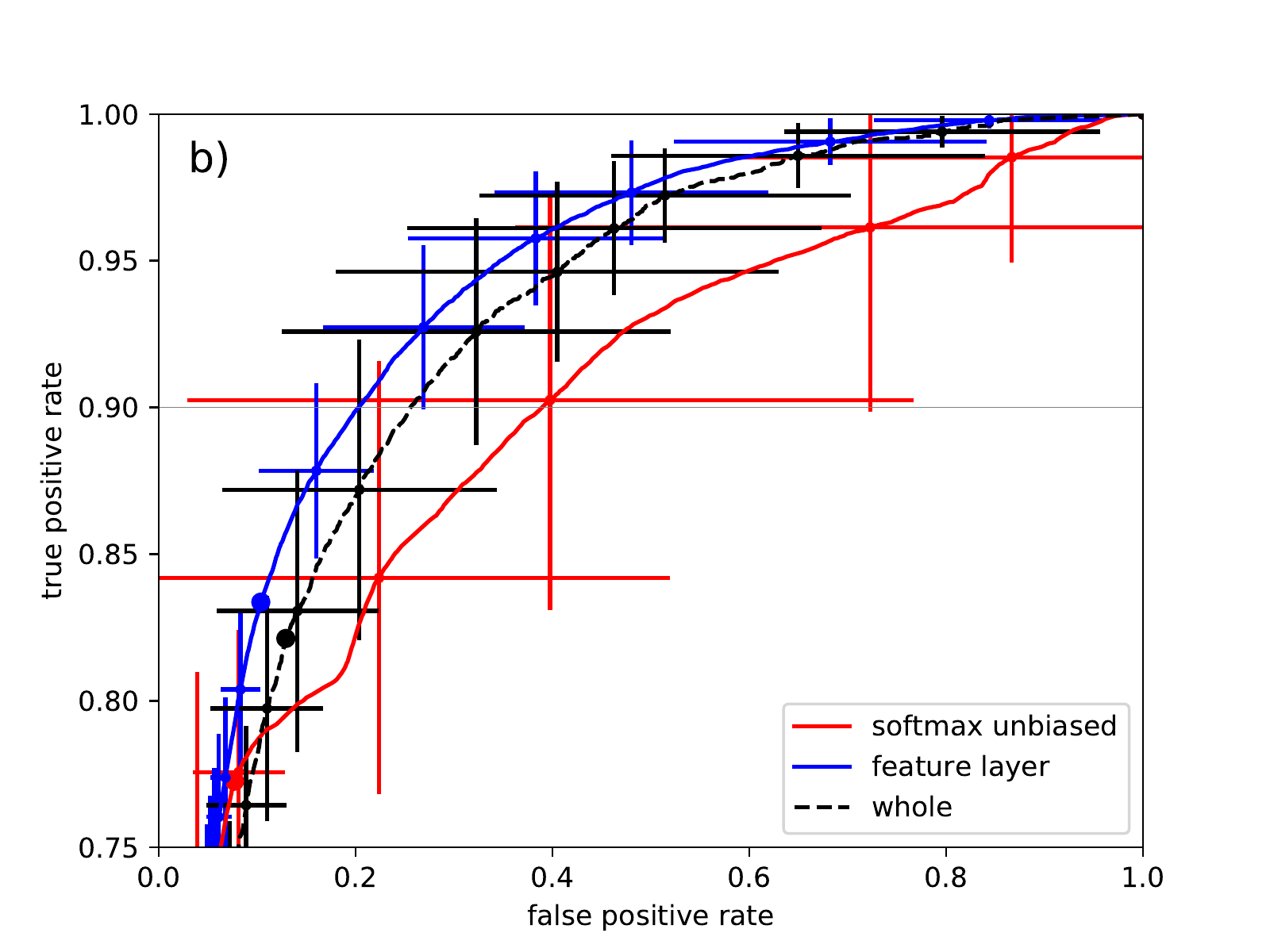}
    \caption{(a) The top quarter of the ROC curve for different input and training of the commitee machine (DenseNet169 backbone). On each curve the point corresponding to the best balanced accuracy is marked by a filled circle, and the 10\% false negative rate is shown by the thin gray line. (b) The variation of the false and true positive rates for a given threshold value, displayed as population standard deviation at selected points.}
    \label{fig:roc}
\end{figure}

\begin{table}[h]
    \centering
    \begin{tabular}{lcccc}

committee machine&ROC AUC&best balanced accuracy&fpr at fnr=0.1\\
\hline
softmax unbiased&$0.9302\pm 0.0077$&$0.8477$&$0.3893$\\
softmax bias=5&$0.9313\pm 0.0101$&$0.8573$&$0.2838$\\
softmax bias=10&$0.9326\pm 0.0092$&$0.8586$&$0.2568$\\
softmax bias=20&$0.9320\pm 0.0095$&$0.8575$&$0.2556$\\
softmax bias=50&$0.9203\pm 0.0513$&$0.8424$&$0.2527$\\
softmax bias=100&$0.7539\pm 0.1763$&$0.6285$&$0.6680$\\
feature layer&$\mathbf{0.9389\pm 0.0105}$&$\mathbf{0.8650}$&$\mathbf{0.2030}$\\
whole&$0.9281\pm 0.0134$&$0.8460$&$0.2556$\\

    \end{tabular}
    \medskip
    \caption{The ROC area under the curve, the best balanced accuracy, and the false positive rate obtained at 10\% false negative rate for different input and training of the committee machine, using the DenseNet169 backbone. The best balanced accuracy and the false positive rate at the selected level are calculated from the averaged ROC curve, see Fig.~\ref{fig:roc}(a).}
    \label{tab:roc}
\end{table}

\begin{table}[t]
    \centering
    \begin{tabular}{lccc}

included feature classifiers&ROC AUC&best balanced accuracy&fpr at fnr=0.1\\
\hline
all &$0.9389\pm 0.0105$&$0.8650$&$0.2030$\\
all except whole&$0.9065\pm 0.0083$&$0.8099$&$0.4165$\\
all except color asymmetry&$0.9387\pm 0.0103$&$0.8655$&$0.1973$\\
all except center&$0.9345\pm 0.0101$&$0.8470$&$0.2779$\\
all except border&$0.9376\pm 0.0102$&$0.8638$&$0.2110$\\
all except blue white veil traits&$0.9389\pm 0.0102$&$0.8651$&$0.2017$\\

    \end{tabular}
    \caption{Performance of the feature layer based committee machine, when one feature classifier is removed from its input (DenseNet196 backbone). }
    \label{tab:ablation}
\end{table}
